\documentclass{article}

\usepackage{arxiv}

\usepackage{amssymb,amsmath,latexsym,amsfonts,amsthm}
\usepackage{mathtools}
\usepackage[utf8]{inputenc} 
\usepackage[T1]{fontenc}    

\usepackage{subcaption}
\usepackage{float}
\usepackage{xcolor}
\usepackage{multirow}
\usepackage{graphicx}

\usepackage[linesnumbered,ruled,vlined]{algorithm2e}
\usepackage{float}
\usepackage{upgreek}

\usepackage{multicol}

\usepackage{caption}
\usepackage{subcaption}

\title{Exploiting epistemic uncertainty of the deep learning models to generate adversarial samples}

\author{
Omer Faruk Tuna \\
  Isik University\\
  Istanbul, Turkey \\
  \texttt{omer.tuna@isikun.edu.tr} \\
  \And
 Ferhat Ozgur Catak \\
  Simula Research Laboratory\\
  Fornebu, Norway \\
  \texttt{ozgur@simula.no} \\
  \And
 M. Taner Eskil \\
  Isik University\\
  Istanbul, Turkey \\
  \texttt{taner.eskil@isikun.edu.tr} 
}

\begin{document}
\maketitle
\begin{abstract}

Deep neural network architectures are considered to be robust to random perturbations. Nevertheless, it was shown that they could be severely vulnerable to slight but carefully crafted perturbations of the input, termed as adversarial samples. In recent years, numerous studies have been conducted in this new area called "Adversarial Machine Learning" to devise new adversarial attacks and to defend against these attacks with more robust DNN architectures. However, almost all the research work so far has been concentrated on utilising model loss function to craft adversarial examples or create robust models. This study explores the usage of quantified epistemic uncertainty obtained from Monte-Carlo Dropout Sampling for adversarial attack purposes by which we perturb the input to the areas where the model has not seen before. We proposed new attack ideas based on the epistemic uncertainty of the model. Our results show that our proposed hybrid attack approach increases the attack success rates from 82.59\% to 85.40\%, 82.86\% to 89.92\% and  88.06\% to 90.03\% on MNIST Digit, MNIST Fashion and CIFAR-10 datasets, respectively. 

\end{abstract}


\section{Introduction}
\label{introduction}
In recent years, deep learning models began to exceed human-level performances. For example, in 2015,  a deep learning model called ResNet \cite{he2015deep} beat the human performance in ImageNet Large Scale Visual Recognition Challenge (ILSVRC) and the record was beaten by more advanced architectures later on. To give some other examples from different benchmark tasks in the image domain, for the problems of reading address information from Google Street View imagery or solving CAPTCHAS, Goodfellow et al. \cite{goodfellow2014multidigit} proposed a system which outperforms human operators. Or, in game playing domain, an AI software named AlphaGo defeated the world Go champion in 2016. Today, we observe that many advanced systems built upon deep learning models offer very high degree of successes in different areas. As a result of this success, deep neural network (DNN) models started to be used in many different fields, ranging from medical diagnosis and autonomous vehicles to game playing or machine translation.  However, during the rise of the DNN models, the researchers' main focus was to build more and more accurate models, and nearly no attention has been paid to the reliability and robustness of these models. Deep learning models indeed require a more elaborate evaluation since these models have some intrinsic vulnerabilities that let intruders easily exploit them. 

By the end of 2013, researchers have discovered that existing deep neural networks are vulnerable to attacks. Szegedy et al. \cite{szegedy2014intriguing} first noticed the presence of adversarial examples in the context image
classification. The authors have shown that it is possible to perturb an image by a small amount and change how the image is classified. Very small and nearly imperceptible perturbations of the data samples are actually sufficient to fool the most advanced classifiers and result in incorrect classification. 
The adversarial machine learning attacks
are based on perturbation of the input instances in a direction that maximizes the chance of wrong decision making and results in false predictions. These attacks can lead to a loss of the model's prediction performance as
the algorithm cannot predict the real output of the input instances correctly.
Thus, attacks utilizing the vulnerability of DNNs can seriously undermine the security of these machine learning (ML) based systems, sometimes with devastating consequences. In the case of medical applications, the perturbation attack can lead to an incorrect diagnosis of a disease. Consequently, it can cause severe harm to the patient's health and also damage the healthcare economy \cite{finlayson2019adversarial}. As another example, autonomous cars use machine learning (ML) to drive traffic without human intervention and avoid accidents. A wrong decision based on an adversarial attack for the autonomous vehicle could cause a fatal accident \cite{sitawarin2018darts,morgulis2019fooling}. Hence, defending against adversarial attempts and increasing the ML models' robustness without compromising clean accuracy is of crucial importance. Assuming that these ML models will serve in critical areas, we should pay the greatest attention to not only ML models' performance but also security concerns of these architectures.

In this study, we focus on adversarial attack strategies based on epistemic uncertainty maximization instead of traditional loss maximization based attacks.  We also look for the most effective approaches that significantly impact their performance and security implications. The adversarial machine learning attacks' current approach is based on model loss maximization and aims to create craftily-designed inputs. Unlike the previous researches in literature, we will follow a slightly modified strategy to craft adversarial samples and try to exploit the model's vulnerability using the model's quantified epistemic uncertainty. We showed that perturbing the input image in a direction that maximizes the model's uncertainty amplifies model loss and results in wrong predictions. The new approach combines adversarial approaches' strengths and weaknesses to produce more destructive attacks by uncertainty and loss maximization.

To sum up; our main contributions for this paper are:

\begin{itemize}
	\item We have utilized a new metric (epistemic uncertainty of the model) which can be exploited to craft adversarial examples. 
	\item We show that the performance of pure uncertainty based attacks is indeed as powerful as the attacks based on the model loss.
	\item We demonstrated that using both the model loss and uncertainty when crafting adversarial examples make up for each other and yields better performance in adversarial attacks.
	
\end{itemize}

This study is organized as follows. Chapter \ref{ch:related_work} will introduce some of the known attack types in literature. In Chapter \ref{ch:preliminaries}, we will introduce the concept of uncertainty together with main types and discuss how we can quantify epistemic uncertainty. Chapter \ref{ch:approach} will give the details of our approach. We will present our experimental results in Section \ref{ch:results} and conclude our work in Section \ref{ch:conclusion}.

\section{Related Work}
\label{ch:related_work}

Since the discovery of DNN's vulnerability to adversarial attacks \cite{szegedy2014intriguing}, a vast amount of research has been conducted in both devising new adversarial attacks and defending against these attacks with more robust DNN models \cite{HUANG2020100270,catak2020generative,9003212,9099439}.

\begin{figure}[!htbp]
    \centering
    \includegraphics[width=0.8\linewidth]{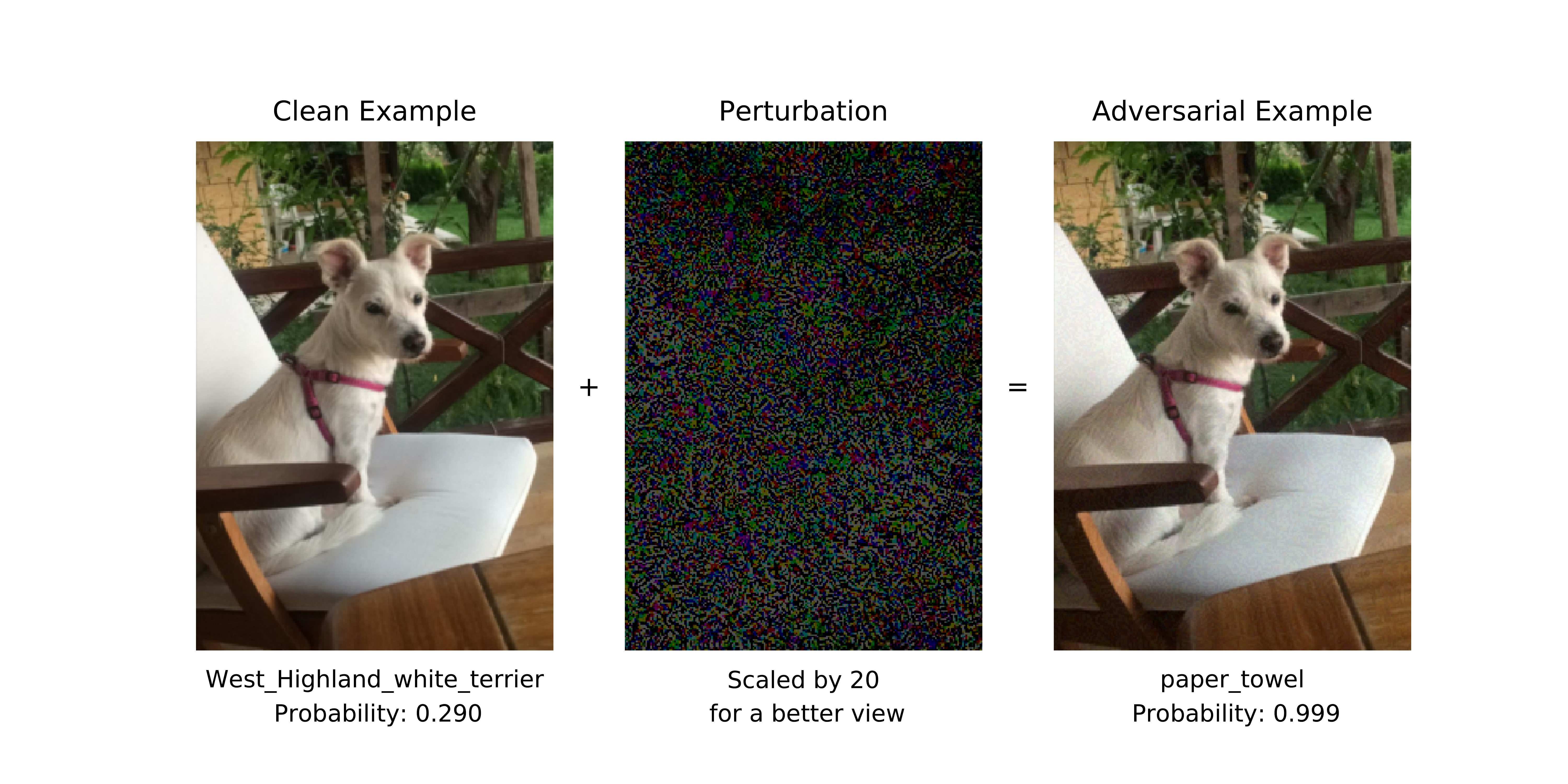}
    \caption{The figure shows an adversarial example. The malicious input is expressed using the original image and the precisely crafted perturbation such that a "West Highland White Terrier (Dog)" is misclassified as "Paper Towel" with high confidence.}
    \label{fig:adv-ml-ex}
\end{figure}

The deep learning models contain many vulnerabilities and weaknesses which make them difficult to defend against in the context of adversarial machine learning. For instance, they are often sensitive to small changes in the input data, resulting in unexpected results in the final output of the model. Figure \ref{fig:adv-ml-ex} shows how an adversary would exploit such a vulnerability and manipulate the model through the use of carefully crafted perturbation applied to the input data.

The attack strategies are mainly based on perturbing the input instance to maximise the model's loss.  An important number of adversarial attack algorithms have been proposed in recent years. The well-knows adversarial attacks are \texttt{Fast-Gradient Sign Method, Iterative Gradient Sign Method, Projected Gradient Descent, Jacobian-Based Saliency Map, Carlini-Wagner}, and \texttt{DeepFool}. Section \ref{sec:fgsm-definition} - \ref{sec:deepfool-definition} describe six different adversarial machine learning attacks briefly.

\subsection{Fast-Gradient Sign Method}\label{sec:fgsm-definition}
This method, also known as FGSM, is one of the earliest and most popular adversarial attacks to date. It is described by Goodfellow et al. \cite{goodfellow2015explaining}. FGSM utilizes the gradient of the model's loss function to adjudge in which direction the pixel values of the source image should be altered to minimize the loss function of the model; and then it changes all pixels simultaneously in the opposite direction to maximize the loss. For a model with classification loss function described as $L(\theta,\mathbf{x},y)$ where $\theta$ represents model parameters, $\mathbf{x}$ is the input to the model (sample input image in our case), $y_{true}$ is the label of our input, we can generate adversarial samples using below formula:

\begin{equation}
    \mathbf{x}^* = \mathbf{x} + \epsilon sign\left(\nabla_x L(\theta,\mathbf{x},y_{true}) \right)
\end{equation}

One last key point about FGSM is that it is not designed to be optimal but fast. That means it is not designed to produce the minimum required adversarial perturbation. Besides, the success ratio of this method is low relatively low in small $\epsilon$ values compared to other attack types.

\subsection{Iterative Gradient Sign Method}
Kurakin et al. \cite{kurakin2017adversarial} proposed a tiny but effective improvement to the FGSM. In their approach, rather than taking only one step of size $\epsilon$ in the direction of the gradient sign, we take several but smaller steps $\alpha$, and we use the given $\epsilon$ value to clip the result. This attack type is often referred to as Basic Iterative Method (BIM), and it is just FGSM applied to an input image iteratively. Generating perturbed images under $L_{inf}$ norm for BIM attack is given by Equation \ref{eq:bim}.

\begin{equation}
\begin{aligned}
\mathbf{x}^* & = \mathbf{x} \\
\mathbf{x}_{N+1}^* & = \mathbf{x} + Clip_{x, \epsilon} \{ \alpha sign \left( \nabla_\mathbf{x} L(\mathbf{x}_N^*, y_{true}) \right) \}
\end{aligned}
\label{eq:bim}
\end{equation}

where $\mathbf{x}$ is the input sample, $\mathbf{x}^*$ is the produced adversarial sample at $i$\textsuperscript{th} iteration, $L$ is the loss function of the model, $y_{true}$ is the actual label for input sample, $\epsilon$ is a tunable parameter, limiting maximum level of perturbation in given $l_{inf}$ norm, $\alpha$ is the step size.

The success ratio of BIM attack is higher than the FGSM  \cite{DBLP:journals/corr/KurakinGB16a}. By adjusting the $\epsilon$ parameter, the attacker can have a chance to manipulate how far past the decision boundary an adversarial sample will be pushed. 

We can group BIM attacks under two main types, namely BIM-A and BIM-B. In the former type, we stop iterations as soon as we succeed in fooling the model (passing the decision boundary), while in the latter, we continue to the attack till the end of provided number of iterations so that we push the input far beyond the decision boundary.

\subsection{Projected Gradient Descent}
This method, also known as PGD, has been introduced by Madry et al. \cite{madry2019deep}. It perturbs a clean image $\mathbf{x}$ for several number of $i$ iterations with a small step size in the direction of model's loss function's gradient. Different from BIM, after each perturbation step, it projects the resulting adversarial sample back onto the $\epsilon$-ball of input sample, instead of clipping. Moreover, instead of starting from the original point ($\epsilon$=0, in all dimensions), PGD uses random start, which can be described as:  

\begin{equation}
    \mathbf{x}_0 = \mathbf{x} + U\left( -\epsilon, +\epsilon \right)
\end{equation}

where $U\left( -\epsilon, +\epsilon \right)$ is the uniform distribution between ($-\epsilon, +\epsilon$).

\subsection{Jacobian-based Saliency Map Attack (JSMA)}

This method, also known as JSMA, has been proposed by Papernot et al. \cite{papernot2017practical}. It is designed to be used under $L_0$ distance norm which takes total number of altered pixels into count when restricting the attacker. It is a greedy algorithm which selects two pixels at a time. The algorithm utilizes the gradient $\nabla Z(x)_l$ to compute a saliency map, which shows the impact of each pixel on the classification of each class. And the aim is to increase the likelihood of target class while decreasing the likelihood of other classes by selecting and updating two pixels at a time based on the saliency map. We continue to the implementation, until either we modify predefined number of pixels or we successfully fool the model.

\subsection{Carlini-Wagner Attack}

This attack type has been introduced by Carlini and Wagner \cite{carlini2017evaluating}, and it is one of the most powerful attack type to date.  Therefore, it is generally used as a benchmark for the adversarial defense research community that aims to create more robust DNN architectures resistant to adversarial attempts.

The authors redefine the attacks as optimization problems which can be solved by using gradient descent to craft more powerful and effective adversarial samples.

\subsection{Deepfool Attack} \label{sec:deepfool-definition}

This attack type has been proposed by Moosavi-Dezfooli et al. \cite{moosavidezfooli2016deepfool} and it is one of the powerful untargeted attack types in literature. It is designed to be used in different distance norm metrics such as $L_{inf}$ and $L_{2}$ norms. 

Deepfool attack has been designed based on an assumption that the neural network models behave as a linear classifier and the classes are separated by a hyperplane. The algorithm starts from the initial input point $\mathbf{x_t}$ and at each iteration it calculates the closest hyperplane and the miminum perturbation amount which is the orthogonal projection to the hyperplane. Then the algorithm calculates $\mathbf{x_t+1}$ by adding the minimal perturbation to the $\mathbf{x}_{t+1}$  and checks whether the misclassification occured. 

\section{Preliminaries} 
\label{ch:preliminaries}

Traditionally, predictive models used to be forced to provide a decision even in ambiguous cases where the model is not sure about it's prediction, and the quality of its predictions is expected to be low. Assuming the model's prediction is always correct without any reasoning by utilizing the model's own uncertainty information may result in catastrophic results. This fact led the researchers to suggest abstaining models based on certain conditions like when the model's uncertainty is high, thus improving the reliability\cite{laves2019uncertainty,tuna2020closeness}.

In this section, we will first introduce the two types of uncertainty in machine learning. And then we will present how we can quantify Epistemic Uncertainty in the context of deep learning.

\subsection{Uncertainty in Machine Learning}\label{sec:preliminaries}

There are two different types of uncertainty in machine learning: epistemic uncertainty and aleatoric uncertainty \cite{hullermeier2020aleatoric,AN2020110617,ZHENG2021107046}.   

\subsubsection{Epistemic Uncertainty}

\begin{figure*}[!htbp]
 \centering
	\includegraphics[width=0.8\linewidth]{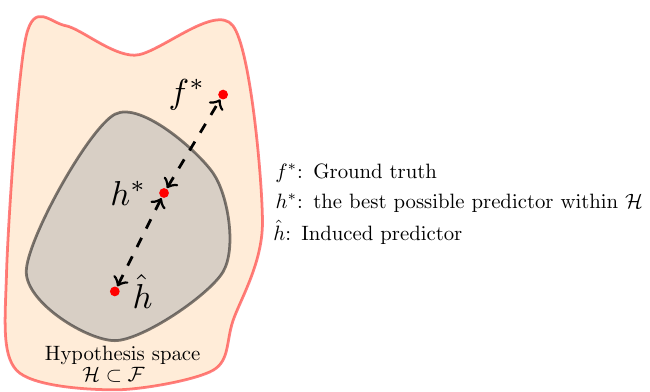}
	\caption{Different types of Epistemic Uncertainty.}
	\label{fig:epistemic-uncertainties}
\end{figure*}

Epistemic uncertainty refers to uncertainty caused by a lack of knowledge and limited data needed for a perfect predictor \cite{ANTONELLI2020746}. It can be categorized under 2 groups as \textit{approximation uncertainty} and \textit{model uncertainty} as depicted in Figure \ref{fig:epistemic-uncertainties}.


\paragraph{Approximation Uncertainty}

In a standard machine learning task, the learner is given a set of data points from an independent, identically distributed dataset. Then he/she tries to induce a hypothesis $\hat{h}$ from the hypothesis space $\mathcal{H}$ by choosing an appropriate learning method with its related hyperparameters and minimizing the expected loss (risk) with a selected loss function, $\ell$. However, what the learner actually does is to try to minimize the \textit{empirical risk} ${R}_{emp}$ which is an estimation of the real risk $R(h)$. The induced $\hat{h}$ is an approximation of the $h^{*}$ which is the optimum hypothesis within $\mathcal{H}$ and the real risk minimizer. This fact results in an approximation uncertainty.  Therefore, the induced hypothesis's quality is not perfect, and the learned model will always be prone to errors.

\paragraph{Model Uncertainty}

Suppose the chosen hypothesis space $\mathcal{H}$ does not include the perfect predictor. In that case, the learner has no chance to realize his/her aim of finding a hypothesis function which can successfully map all possible inputs to outputs. This leads to a discrepancy between the ground truth $f^{*}$ and best possible function $h^{*}$ within $\mathcal{H}$, called model uncertainty. 

However, Universal Approximation Theorem states that for any target function $f$, there actually exists a neural network which can approximate $f$  \cite{zhou2018universality,Cybenko}. The hypothesis space $\mathcal{H}$ is huge for deep neural networks, thus it will not be wrong to assume that $h^{*} = f^{*}$. We can ignore the model uncertainty for deep neural networks, and we can only care about the approximation uncertainty. Consequently, in deep leaning tasks, the actual source of epistemic uncertainty is related to approximation uncertainty. Epistemic uncertainty refers to the confidence a model has about its prediction \cite{Loquercio_2020}. The underlying cause is the uncertainty about the parameters of the model. This type of uncertainty is apparent in the areas where we have limited training data, and the model weights are not optimized correctly.

\subsubsection{Aleatoric Uncertainty}

Aleatoric uncertainty refers to the variability in an experiment's outcome, which is due to the inherent random effects \cite{GUREVICH2019291}. This type of uncertainty can not be reduced even if we have enough number of training samples \cite{SENGE201416}. An excellent example of this phenomenon is the noise observed in the measurements of a sensor. 


\begin{figure}[!htbp]
 \centering
	\includegraphics[width=0.8\linewidth]{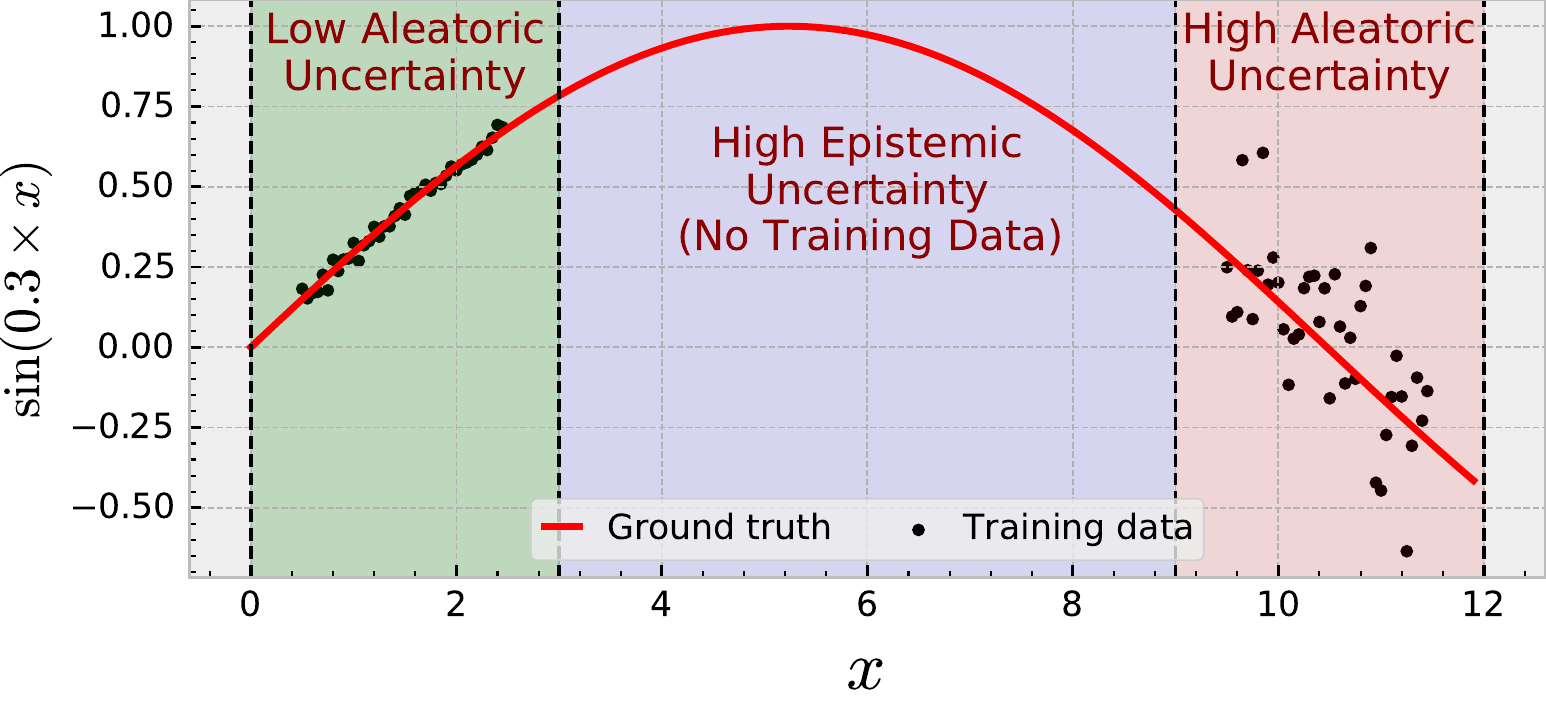}
	\caption{Illustration of the Epistemic and Aleatoric uncertainty.}
	\label{fig:uncertainties_lat}
\end{figure}

Figure \ref{fig:uncertainties_lat} shows a simple nonlinear function ( $\sin({0.3 \times x})$ where $x\in[0,12]$ ) plot. As shown in the region where data points are populated at right ($9<x<12$), the noisy samples are clustered, leading to high aleatoric uncertainty. As an example, these points may represent measurements of a faulty sensor; one can conclude that the sensor produces errors around $x=10.5$ by some inherent reason. We can also conclude that the middle regions of the figure represent the high epistemic uncertainty areas. Because there are not enough training samples for our model to describe the data best. Moreover, we can claim that the high epistemic uncertainty area represents the low prediction accuracy area.

\subsection{Quantifying Epistemic Uncertainty in Deep Neural Networks}\label{sec:preliminaries}

Using techniques which helps us to quantify the uncertainty of the model is necessary for healthy decision making. Assuming that we, as humanity, will use deep learning models in the areas where safety and reliability is a critical concern such as autonomous driving, medical applications; researchers need to be very careful and pay utmost attention to prediction uncertainty. This will obviously help us to increase the quality of the predictions.

In the recent years, there have been significant number of researches conducted to quantify uncertainty in deep learning models. Most of the work was based on Bayesian Neural Networks which learn the posterior distribution over weights to quantify predictive uncertainty \cite{Hinton1995BayesianLF}. However, the Bayesian NN's come with additional computational cost and inference issue. Therefore, several approximations to Bayesian methods have been developed which make use of variational inference \cite{NIPS2011_7eb3c8be,paisley2012variational,hoffman2013stochastic,blundell2015weight}. On the other hand, Lakshminarayanan et al. \cite{lakshminarayanan2017simple} used deep ensemble approach as an alternative to Bayesian NN's to quantify predictive uncertainty. But this requires training several NN's which my not be feasible in practice. A more efficient and elegant approach was proposed by Gal et al. \cite{gal2016dropout}. The authors showed that a neural network model with inference time dropout is equivalent to a Bayesian approximation of the Gaussian process. And the prediction hypothesis uncertainty can be approximated by averaging probabilistic feed-forward Monte Carlo dropout sampling during the prediction time. 

It acts as an ensemble approach. In each single ensemble model, the system has to drop out different neurons in the network's each layer according to the dropout ratio in the prediction time. The predictive mean is the average of the predictions over dropout iterations, $T$, and the predictive mean is used as the final inference $\hat{y}$, for the input instance $\mathbf{\hat{x}}$ in the dataset. The overall prediction uncertainty is approximated by finding the variance of the probabilistic feed-forward Monte Carlo (MC) dropout sampling during prediction time. The prediction is defined as follows:

\begin{equation}
 p(\hat{y}=c|\mathbf{\hat{x}},\mathcal{D}) \approx \hat{\mu}_{pred}  = \frac{1}{T} \sum_{y \in T} p(\hat{y}|\mathbf{\theta},\mathcal{D})
\end{equation}

where $\theta$ is the model weights, $\mathcal{D}$ is the input dataset, $T$ is the number of predictions of the MC dropouts, and $\mathbf{x}$ is the input sample. The label of input sample $\mathbf{x}$ can be estimated with the mean value of Monte-Carlo dropouts predictions $p(\hat{y}|\mathbf{\theta},\mathcal{D})$, which will be done $T$ times.

\begin{figure}[!htbp]
	\includegraphics[width=1.0\linewidth]{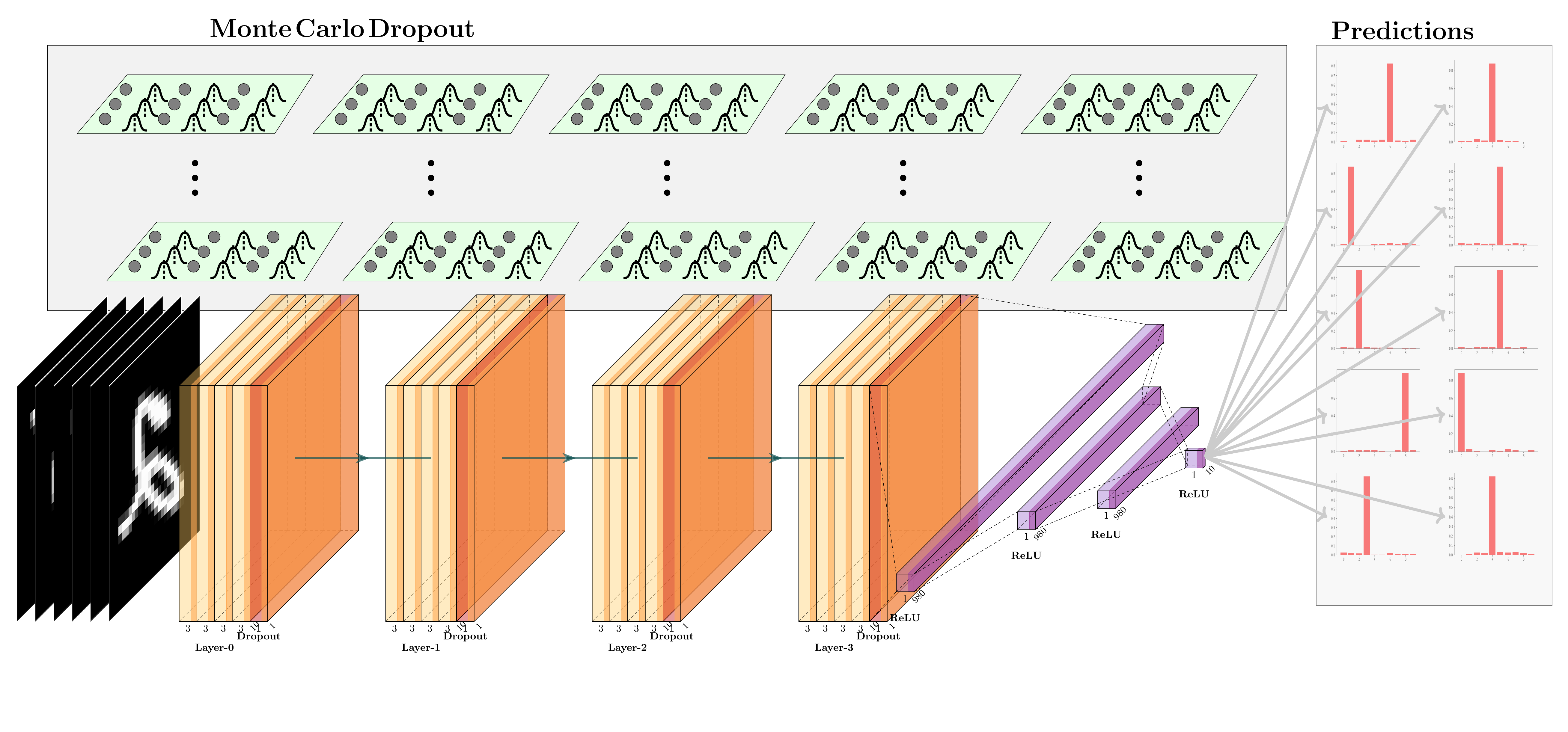}
	\caption{Illustration of the Monte Carlo dropout based Bayesian prediction.}
	\label{fig:mc-dropouts}
\end{figure}

Figure \ref{fig:mc-dropouts} shows the general overview of the Monte Carlo dropout based classification algorithm. In the prediction time, random neurons in each layer are dropped out (based on the probability $p$) from the base neural network model to create a new model. As a result, $T$ different classification models can be used for the prediction of the input instance's class label and uncertainty quantification of the overall prediction. For each testing input sample $\mathbf{x}$, the predicted label is assigned with the highest predictive mean. And the variance of the $p(\hat{y})$ is can be used as a measure of epistemic uncertainty of the model.


We have chosen MC dropout method due to its simplicity and efficiency. It is computationally much more efficient than the other approaches. The approach needs only a single trained model to measure the uncertainty, while the different techniques such as DeepEnsemble need multiple models. Secondly, one can take the backward derivative of the computed variance term for each input sample and use it to craft adversarial examples to evade the model. 

\section{Approach}\label{ch:approach}

\begin{figure*}[!htbp]
 \centering
	\includegraphics[width=0.8\linewidth]{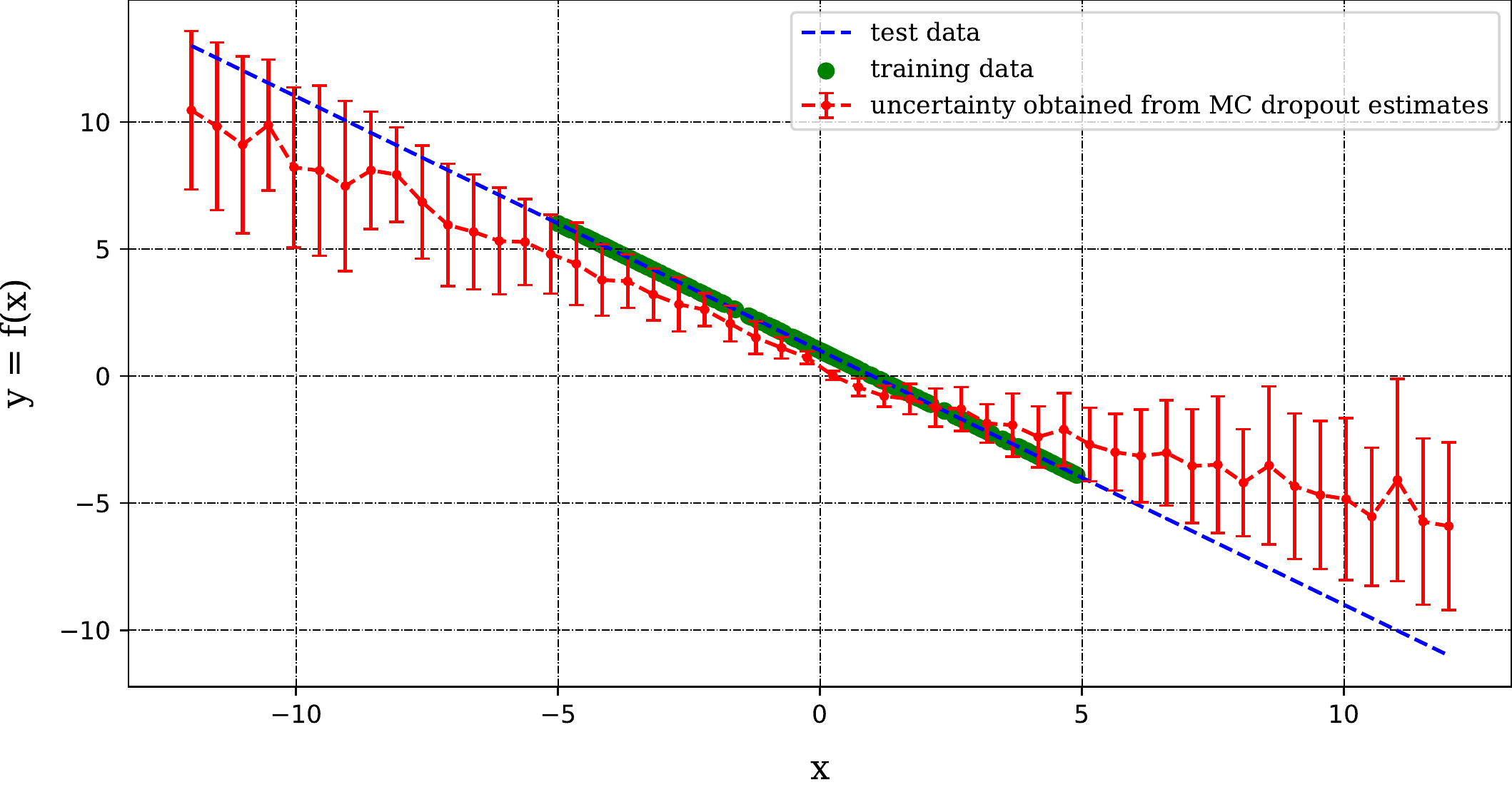}
	\caption{Uncertainty values obtained from a regression model}
	\label{fig:regression}
\end{figure*}

It is obvious that model uncertainty is higher in the areas where we have a limited number of training points. Due to this ignorance about ground truth, we cannot achieve a perfect model that can predict accurately every possible test data. Figure \ref{fig:regression} shows a regression model's prediction outputs trained on a limited number of data points constrained on some interval. In this simple example, we trained a single hidden layer NN with ten neurons to learn a linear function $y = -x + 1$. As can be seen from the graph, in the areas where we do not have enough training points, the uncertainty values obtained from MC dropout estimates of our model is high, which can be interpreted as the quality of the prediction is low, and our model is having difficulties in deciding the correct output values. Harmoniously, we also observe high loss values in these areas. For this reason, we can conclude that the high epistemic uncertainty area represents the low prediction accuracy area. Accordingly, we claim that pushing the model's limits by testing it in extreme conditions with input that our model has never seen before may result in failure of model prediction output.

The aim of the adversarial attacks is to find a least perturbation amount ($\delta$) constrained to some interval ($\epsilon$), resulting in maximum loss, thus fooling the classifier. We can express this mathematically in the below equation, where  $F_\theta(x)$ is our neural network.

\begin{equation}
\underset{\|\delta\| \leq \epsilon}{arg\,max} \ell(F_\theta(x), y)
\end{equation}

Like most of the attack types in literature, the attackers perturb the input image in a direction which maximizes the loss, and this direction is found using the gradient of the loss function. However, we showed that instead of using the loss function, another effective approach is to use the model's epistemic uncertainty. Our alternative approach uses the model's epistemic uncertainty as a tool for creating successfully manipulated adversarial input instances. In contrast to the loss based adversarial machine learning attacks,  this method can provide an alternative strategy in which the attacker can make an effective attack (by generating a specific error in a particular uncertainty direction). 

\begin{figure}[!htbp]
    \centering
        \begin{subfigure}[b]{0.33\linewidth}
         \centering
         \includegraphics[width=1\linewidth]{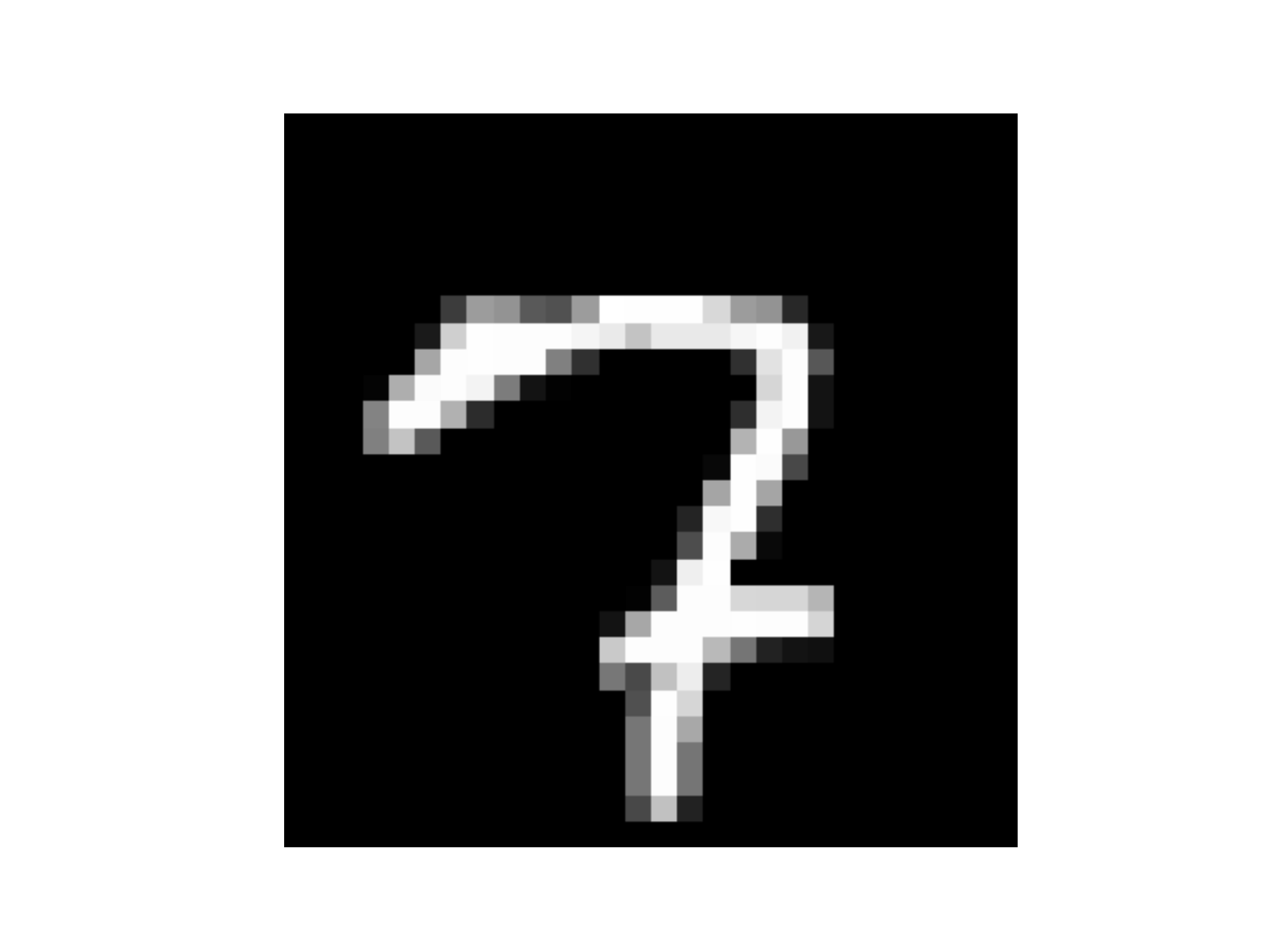}
         \caption{MNIST Digit test image number 37}
	    \label{fig:img-1}
     \end{subfigure}\hfill
    \begin{subfigure}[b]{0.33\linewidth}
         \centering
         \includegraphics[width=0.9\linewidth]{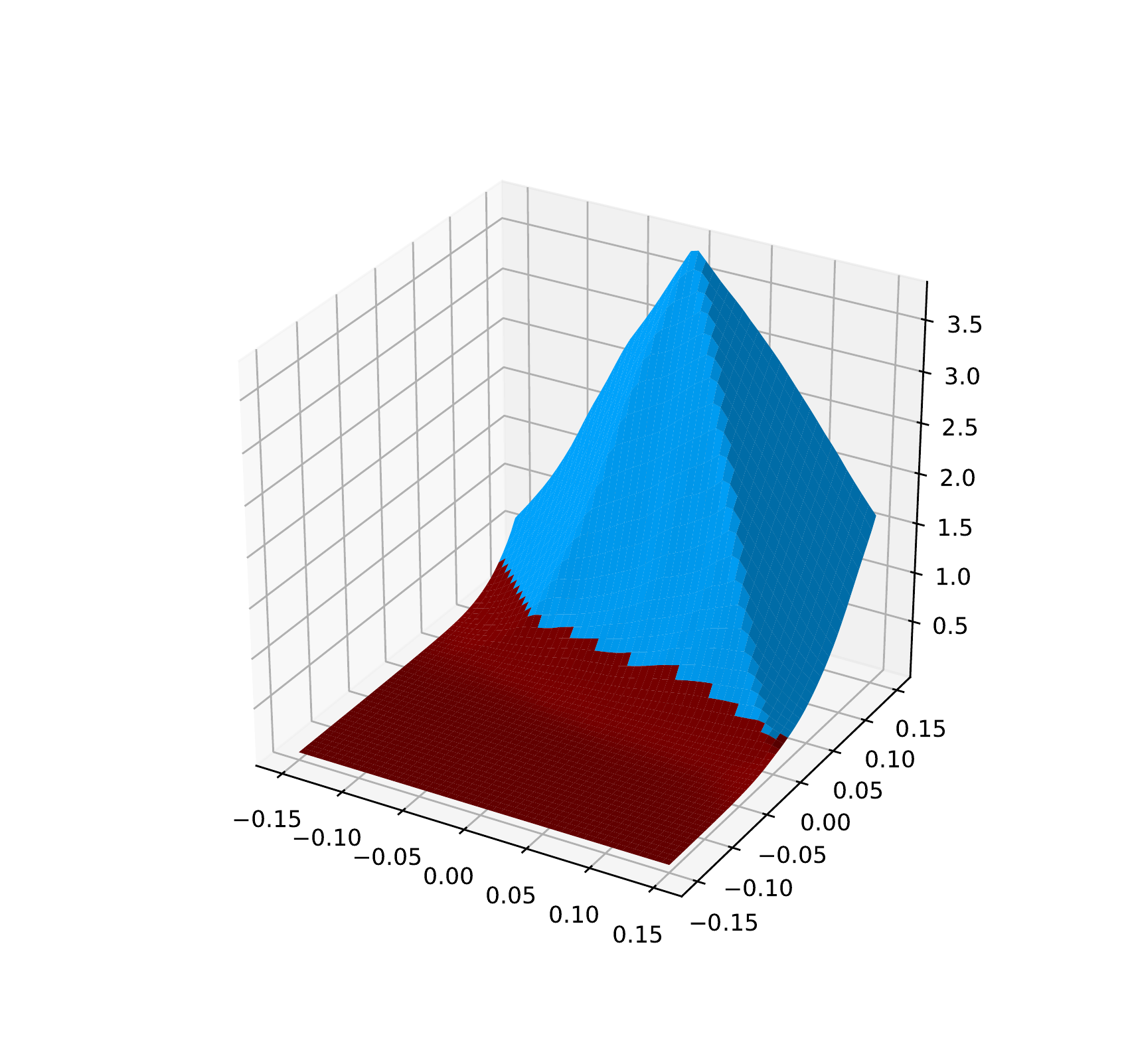}
         \caption{Loss gradient direction}
	    \label{fig:loss-surface-in-loss-direction}
     \end{subfigure}\hfill
     \begin{subfigure}[b]{0.33\linewidth}
         \centering
         \includegraphics[width=0.9\linewidth]{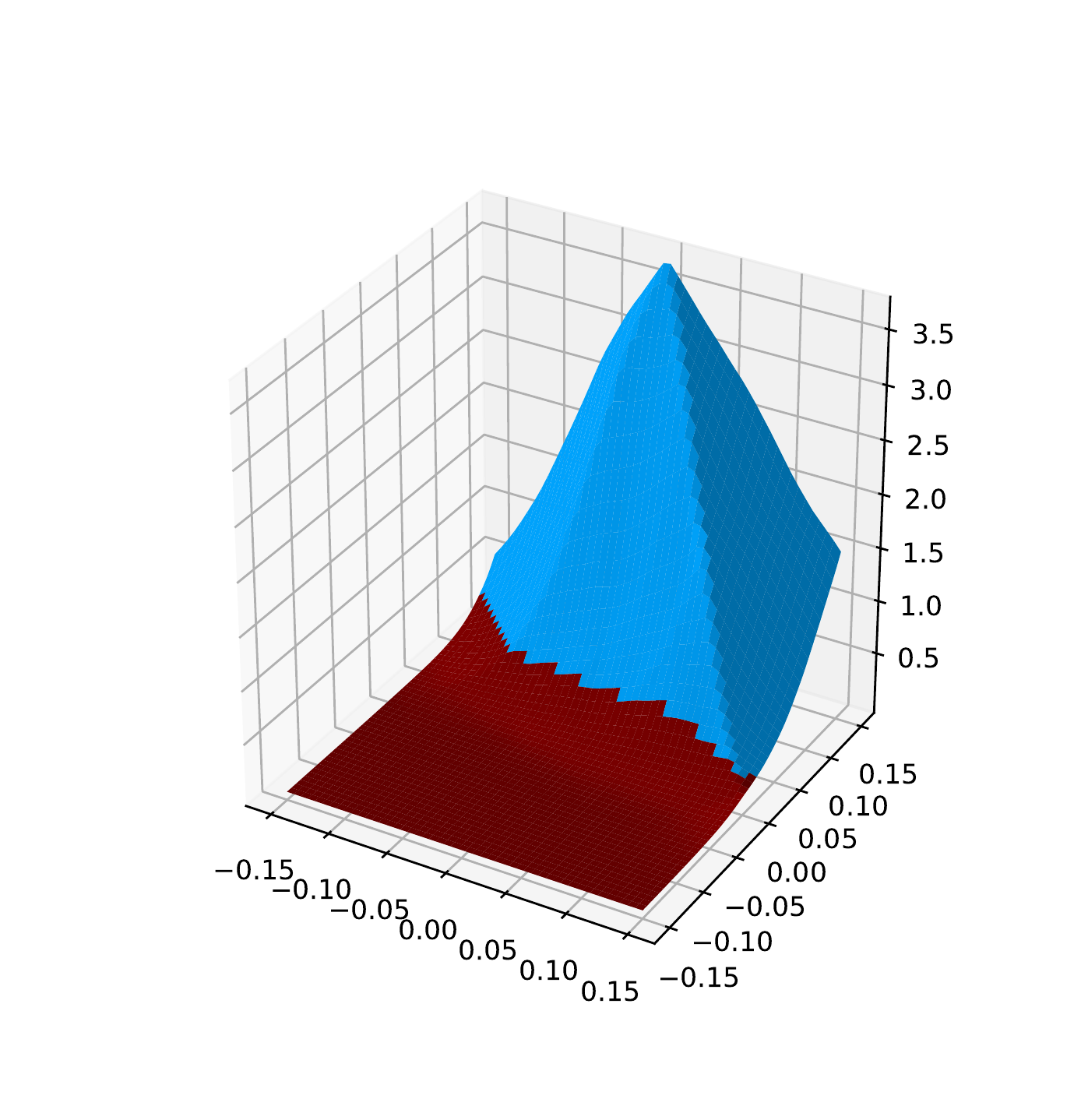}
         \caption{Uncertainty gradient direction}
	    \label{fig:loss-surface-in-uncertainty-direction}
     \end{subfigure}\hfill
     \begin{subfigure}[b]{0.33\linewidth}
         \centering
         \includegraphics[width=0.9\linewidth]{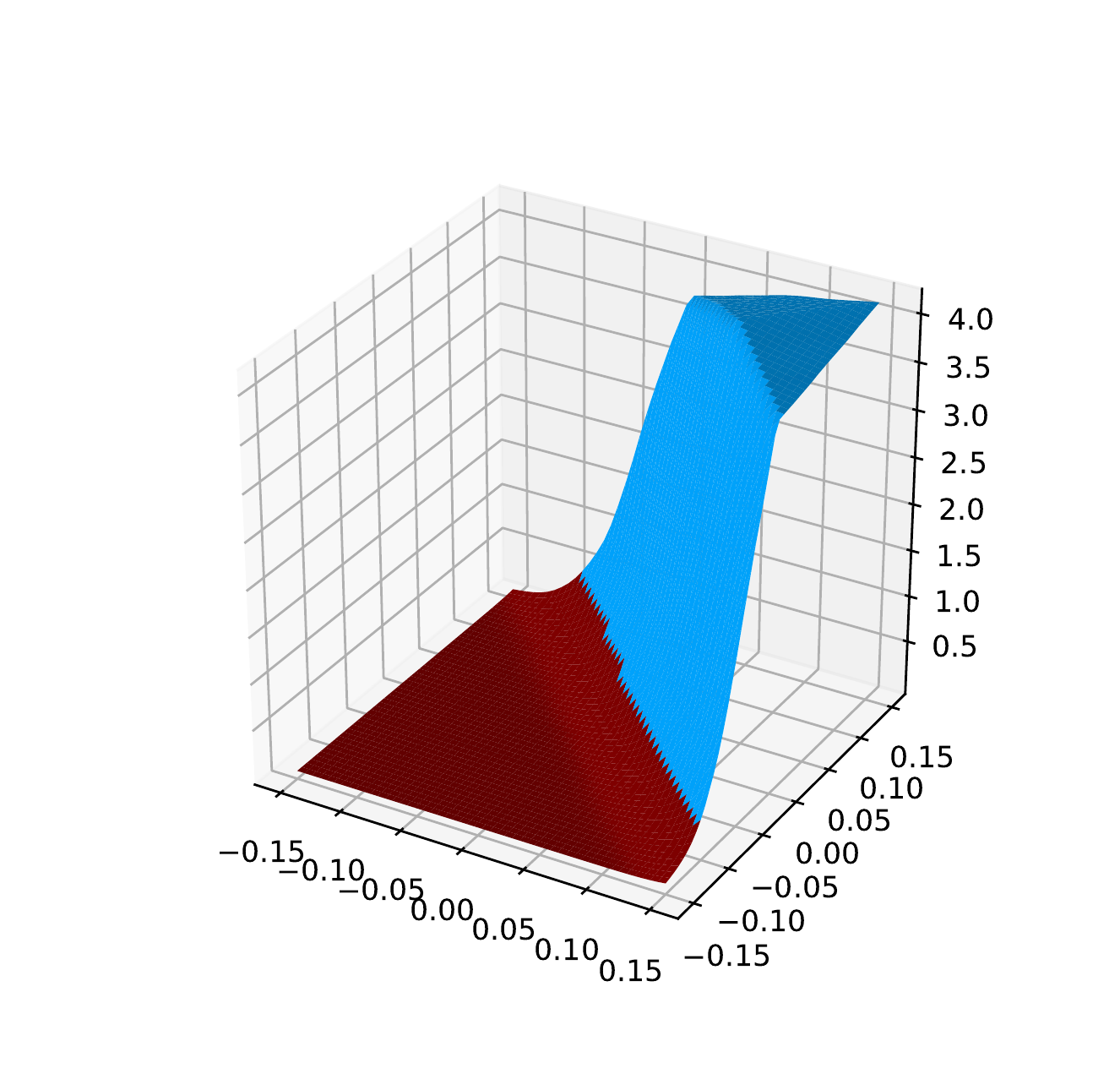}
         \caption{Hybrid gradient direction}
	    \label{fig:loss-surface-in-hybrid-direction}
     \end{subfigure}\hfill
	\caption{Loss surfaces in different directions within constrained $\epsilon$ neighborhood. The maximum loss values are b) 3.783, c) 3.713, d) 4.167}
	\label{fig:loss-surfaces}
\end{figure}

To verify that our intuition holds true, we have made a simple experiment and depicted the loss surfaces of a trained CNN model (digit classifier) within a constrained epsilon neighbourhood of the original input data points as can be seen in Figure \ref{fig:loss-surfaces}. Figure \ref{fig:loss-surface-in-loss-direction} shows the model's loss values in the direction of the model's loss gradient and a random direction. We see that the maximum loss value observed is 3.783. Then, as shown in Figure \ref{fig:loss-surface-in-uncertainty-direction}, we projected the model loss surface to the direction of the model's epistemic uncertainty's gradient and the same random direction we used in the previous try. This time, the maximum loss value achieved is 3.713, which is very close to the previous one. When we analyzed the directions of loss' and uncertainty's gradients, we saw that out of 784 sub directions, 693 of them were the same and 91 of them were different. It means that we can maximize the model loss by perturbing the input image in a slightly different direction than we used to do before. In our last attempt, we depicted the model loss surface in the direction of loss and uncertainty's gradient directions as in \ref{fig:loss-surface-in-hybrid-direction}. We could now reach a loss value of 4.167, which is bigger than the previous two attempts. In Figures \ref{fig:loss-surface-in-loss-direction},\ref{fig:loss-surface-in-uncertainty-direction}, the points where we see a difference in color on the loss surfaces indicate that the model prediction has changed from the correct class "7" to wrong class "2". Therefore, we can conclude that perturbing the image in both directions will lead to misclassification for the model. 

It is a well-known point that the loss curve of a deep neural network with a high nonlinearity level has lots of local minimums and maximums in a high dimensional space. Numerical solution to finding global extrema points is an NP-hard problem \cite{10.5555/78221,BLUM1992117}. No optimization approach can reach these global extrema points by utilizing a naive method like gradient descent on the model's loss function. Eventually, the optimizer will be stuck to local extrema points. However, the above experiment showed us a hint that slightly changing the direction in each gradient descent step by leveraging the model uncertainty can increase the proposed attacks' performance.

\begin{figure}[!htbp]
    \centering
            \begin{subfigure}[b]{0.33\linewidth}
         \centering
         \includegraphics[width=1\linewidth]{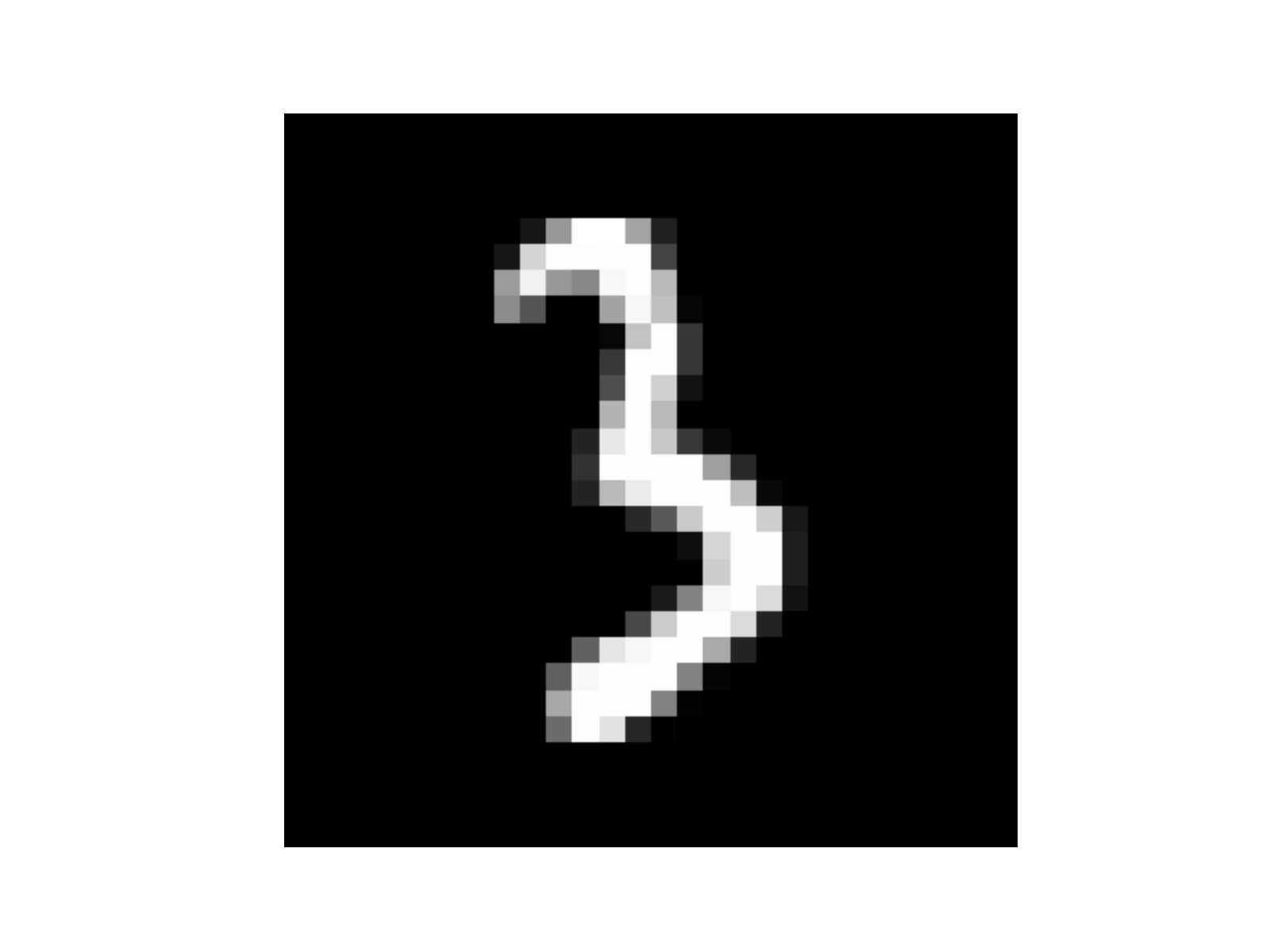}
         \caption{MNIST Digit test image number 45}
	    \label{fig:img-2}
     \end{subfigure}\hfill
    \begin{subfigure}[b]{0.33\linewidth}
         \centering
         \includegraphics[width=1\linewidth]{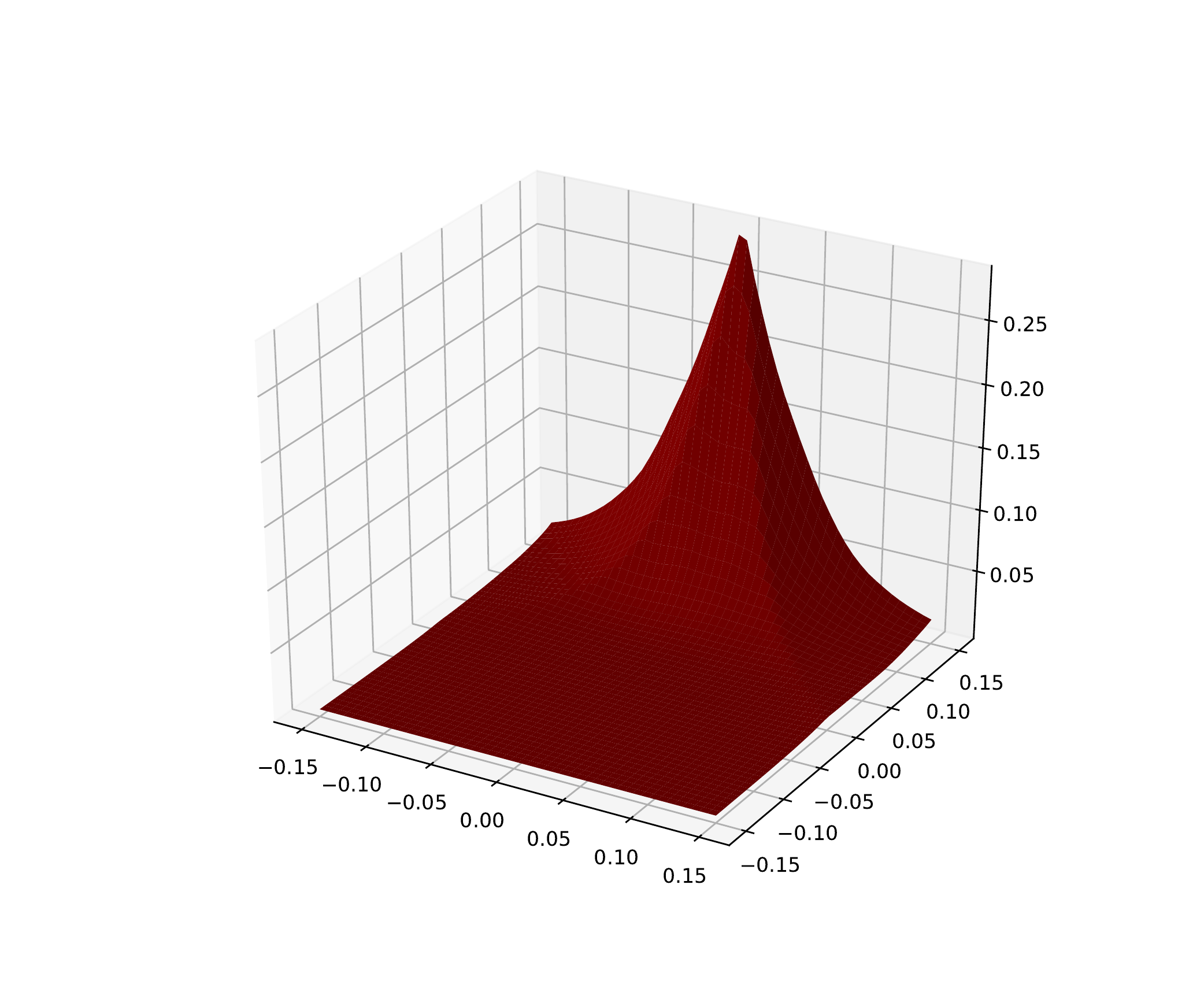}
         \caption{Loss gradient direction}
	    \label{fig:loss-surface-in-loss-direction-2}
     \end{subfigure}\hfill
     \begin{subfigure}[b]{0.33\linewidth}
         \centering
         \includegraphics[width=0.9\linewidth]{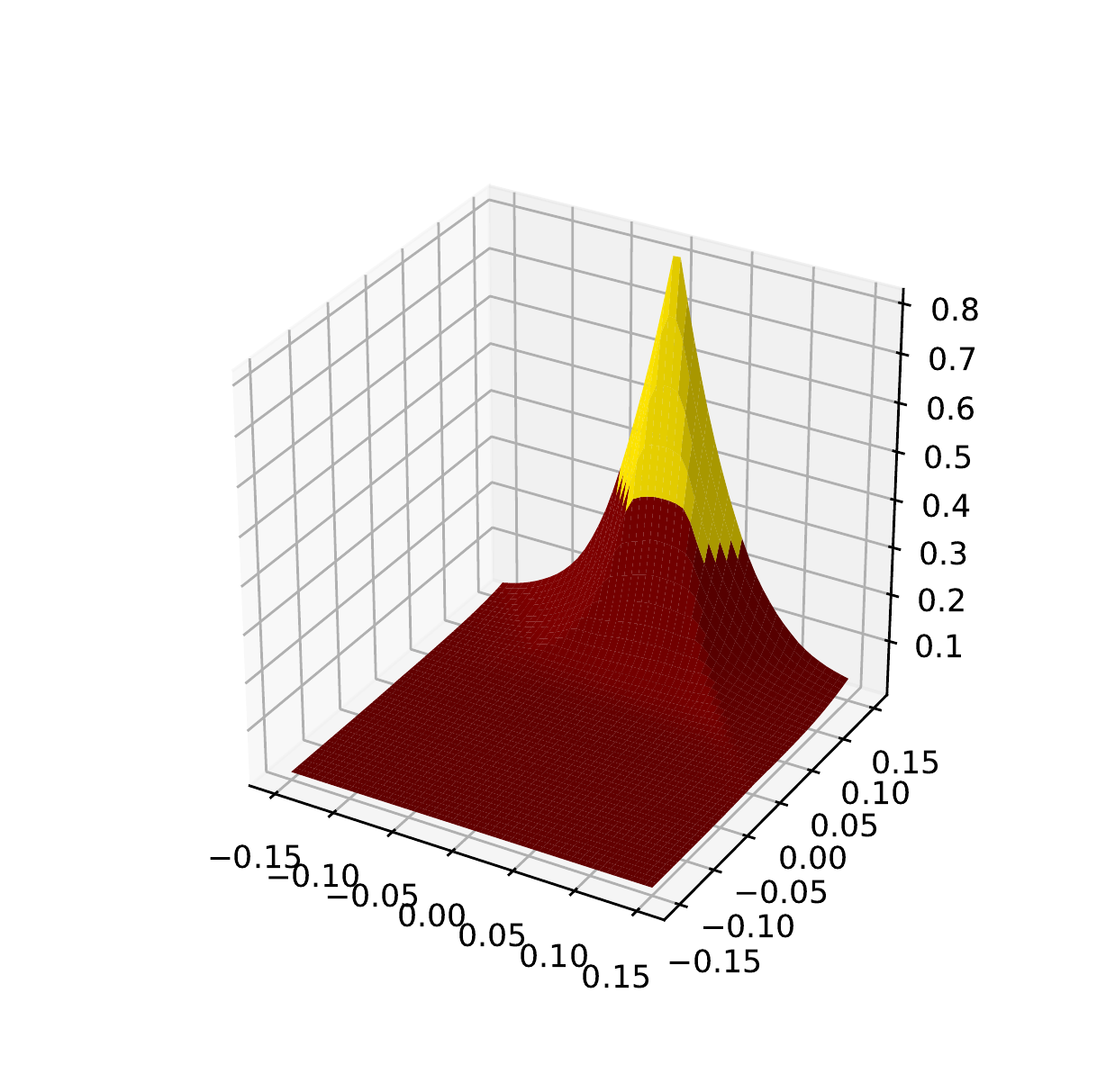}
         \caption{Uncertainty gradient direction}
	    \label{fig:loss-surface-in-uncertainty-direction-2}
     \end{subfigure}\hfill
     \begin{subfigure}[b]{0.33\linewidth}
         \centering
         \includegraphics[width=0.9\linewidth]{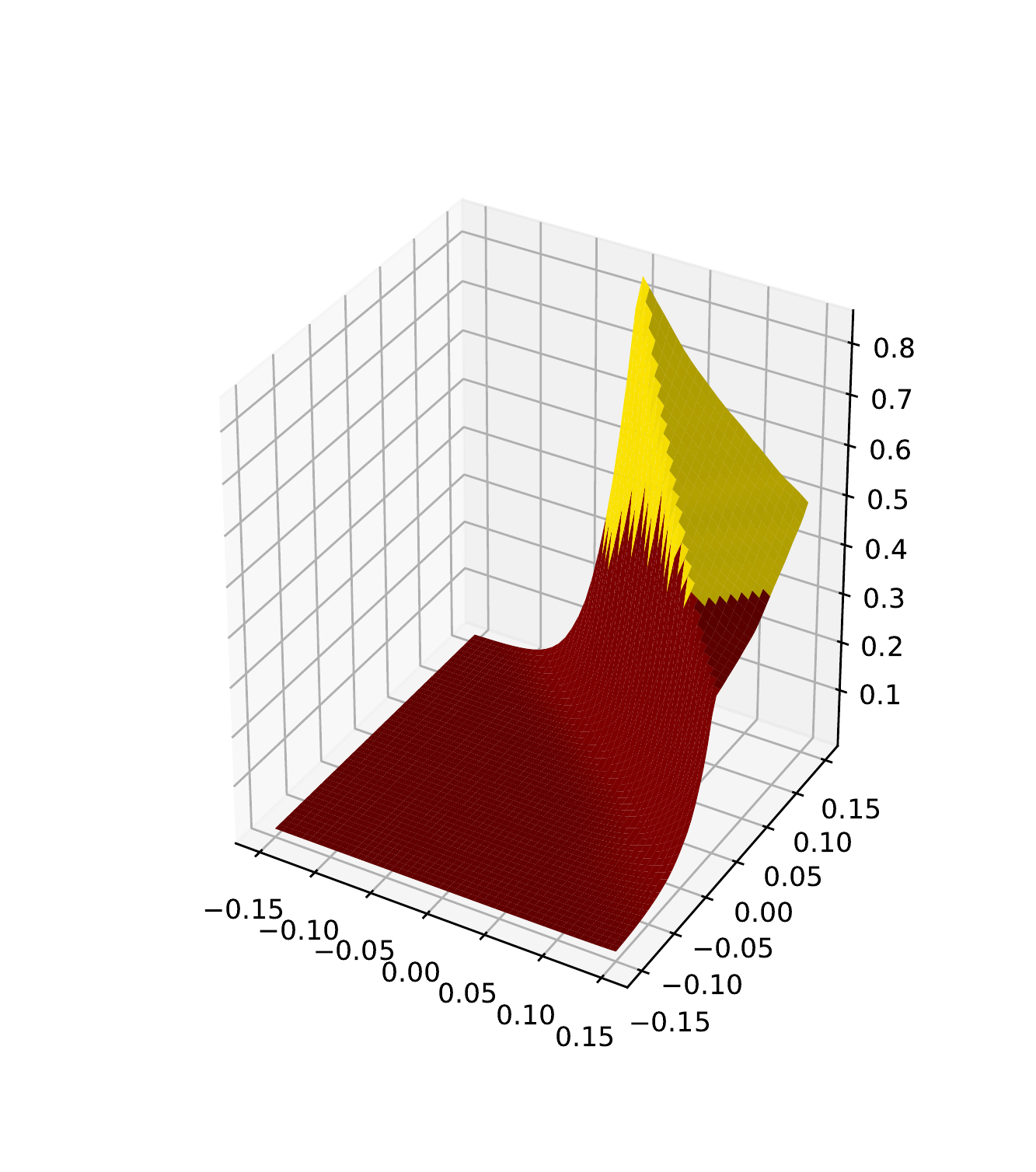}
         \caption{Hybrid gradient direction}
	    \label{fig:loss-surface-in-hybrid-direction-2}
     \end{subfigure}\hfill
	\caption{Loss surfaces in different directions within constrained $\epsilon$ neighborhood.  The maximum loss values are b) 0.285, c) 0.811, d) 0.845}
	\label{fig:loss-surfaces-2}
\end{figure}

We conducted the same experiment on a different sample from the MNIST test dataset. Figure \ref{fig:loss-surfaces-2} shows that the maximum loss value in uncertainty is far greater than the maximum loss value in model loss' gradient direction. And the maximum loss value in the hybrid direction is larger than the ones in both model loss' and uncertainty's directions. Besides, we observe that there is no possibility of misclassification in the loss gradient direction, as there is no visible colour change in the surface plot of Figure \ref{fig:loss-surface-in-loss-direction-2}, whereas in Figure \ref{fig:loss-surface-in-uncertainty-direction-2} we observe that there are yellow regions where the model misclassifies the input image in the uncertainty's gradient direction. Again, when we analyzed the directions of loss' and uncertainty's gradients, we saw that out of 784 directions, only 639 of them were the same and 145 of them were different which is much larger than the first experiment.

The epistemic uncertainty yields a better direction for our second experiment because our model (like all the trained ML models) is not the "perfect" predictor and is just an approximation to the oracle function. The model itself has an inherent "approximation uncertainty" which sometimes induce to sub-optimal solutions. Consequently, any method which only relies on the trained model (which is not the optimum model) will result in less effective performance.

\subsection{Proposed Epistemic Uncertainty Based Attacks}\label{sec:approach}
\label{sec:approach}

Previous attack types in literature have been designed to exploit the model loss and aimed at maximizing the model loss value within a constrained neighbourhood of the input data points. And we have witnessed quite successful results with this approach. However, one possible drawback for these kinds of attacks is that they solely rely on the trained ML model, which inevitably suffers from the approximation error. We can overcome this problem by utilizing an additional metric, namely epistemic uncertainty of the model. This additional uncertainty information has a correcting effect and improves the convergence to global extrema points by yielding higher loss value. Results shown in Figure  \ref{fig:loss-surfaces} and Figure \ref{fig:loss-surfaces-2} supports our argument. Therefore, we can reformulate existing attacks using model $uncertainty$ instead of model $loss$. And even we can benefit from both of them.

\subsubsection{Fast Gradient Sign Method}

\begin{equation}
    \mathbf{x}^{adv} = \mathbf{x} + \epsilon \cdot sign(\nabla_{\mathbf{x}} \ell(\mathbf{x},y_{true}))
\end{equation}
where $\mathbf{x}$ is the input (clean) image $x^{adv}$ is the perturbed adversarial image, $\ell$ is the classification loss function, $y_{true}$ is true label for the input $\mathbf{x}$.

Our uncertainty based modified FGSM attack is shown as;

\begin{equation}
    \mathbf{x}^{adv} = \mathbf{x} + \epsilon \cdot sign(\nabla_{\mathbf{x}} U(\mathbf{x}, F, p, T))
\end{equation}
where $\mathbf{x}$ is the input (clean) image $\mathbf{x}^{adv}$ is the perturbed adversarial image $U$ is the uncertainty metric (mean variance) obtained from $T$ different MC dropout estimates, $F$ is the prediction model in training mode, $p$ is the dropout ratio used in the dropout layers, $T$ is the number of MC dropout samples in model training mode.

Calculation for Uncertainty metric U (mean variance of T predictions):

\begin{itemize}
    \item[\textbf{Step: 1}] For an input image $\mathbf{x}$, $T$ different predictions is obtained $p_t(\mathbf{x})$ by Monte-Carlo Dropout sampling where each prediction is a vector of softmax scores for the $C$ classes. $$p_t (\mathbf{x})= \mathcal{F}(\mathbf{x},p,T)$$ Where $\mathcal{F}$ is the prediction model in training mode, $p$ is the dropout ratio used in the dropout layers, $T$ is the number of MC dropout samples in model training mode 
    \item[\textbf{Step 2:}] The next step is to compute the average prediction score for the $T$ different outputs: $$p_T (\mathbf{x})=  \frac{1}{T} \sum_{t \in T}p_t(\mathbf{x}) $$ 
    \item[\textbf{Step 3:}] Compute the variance of the $T$ predictions for each class. $$\sigma^2 (p_t(\mathbf{x})) = \frac{1}{T} \sum_{t \in T} (p_t(\mathbf{x}) - p_T(\mathbf{x}) )^2 $$
    \item[\textbf{Step 4:}] Compute the expected value of variance over all classes by taking their average. $$ U(\mathbf{x},F,p,T)=E(\sigma^2 (p_T (\mathbf{x})))$$
\end{itemize}

\subsubsection{Basic Iterative Attack (Uncertainty Based)}

In this part, we first provide the pseudo-codes for known Loss Based BIM attack types as in Algorithm \ref{alg:BIM-A-Loss-Alg} and \ref{alg:BIM-B-Loss-Alg}. Then, we provide our proposed uncertainty based BIM attack variants in Algorithm \ref{alg:BIM-A-Uncertainty-Alg} and \ref{alg:BIM-B-Uncertainty-Alg}. All the attack types proposed here are designed under $L_\infty$ norm.

        \begin{algorithm}[H]
        \DontPrintSemicolon
          \KwIn{$\mathbf{x} \in \mathbb{R}^m, y_{true}, F, N, \epsilon, \alpha$}
          \KwOut{$\mathbf{x}_{t+1}$ }
          $\mathbf{x}_0 \gets \mathbf{x} $ \;
          \While{$n < N$}
          {
            \tcc{update X by using below formula, F in evaluation mode}
            $\mathbf{x}_{(t+1)} = clip_{\mathbf{x}, \epsilon}(\mathbf{x}_t + \alpha \cdot sign(\nabla_\mathbf{x} \ell(\mathbf{x}_t, y_{true}))) $ \;
            \If{$arg\,max ( F(\mathbf{x}_{t+1}) ) \neq y_{true} $}
            {
                end while\;
            }
          }
          $return \,\, \mathbf{x}_{t+1}$
        \caption{Algorithm for BIM A (Loss based) \\ $\mathbf{x}$ is the benign image, $y_{true}$ is the true label for $\mathbf{x}$, $F$ is the model function learnt during training, $N$ is the number of iterations, $\epsilon$ is the maximum allowed perturbation, $\alpha$ is the step size.}
        \label{alg:BIM-A-Loss-Alg}
        \end{algorithm}

        \begin{algorithm}[H]
        \DontPrintSemicolon
          \KwIn{ $\mathbf{x}  \in \mathbb{R}^m, y_{true}, F, N, \epsilon, \alpha$}
          \KwOut{$\mathbf{x}_{t+1}$ }
          $\mathbf{x}_0 \gets \mathbf{x} $ \;
          \While{$n < N$}
          {
            \tcc{update X by using below formula, F in evaluation mode}
            $\mathbf{x}_{(t+1)} = clip_{\mathbf{x}, \epsilon}(\mathbf{x}_t + \alpha \cdot sign(\nabla_\mathbf{x} \ell(\mathbf{x}_t, y_{true}))) $ \;
          }
          $return \,\, \mathbf{x}_{t+1}$
        \caption{Algorithm for BIM B (Loss based) \\ $\mathbf{x}$ is the benign image, $y_{true}$ is the true label for $\mathbf{x}$, $F$ is the function learned by the network during training, $N$ is the number of iterations, $\epsilon$ is the maximum allowed perturbation, $\alpha$ is the step size.}
        \label{alg:BIM-B-Loss-Alg}
        \end{algorithm}

        \begin{algorithm}[H]
        \DontPrintSemicolon
          \KwIn{$\mathbf{x}  \in \mathbb{R}^m, F, p, T, N, \epsilon, \alpha$}
          \KwOut{$\mathbf{x}_{t+1}$ }
          $\mathbf{x}_0 \gets \mathbf{x} $ \;
          $initial prediction = arg\,max(F(\mathbf{x}_0))$ \; 
          \While{$n < N$}
          {
            \tcc{update X by using below formula, F in training mode}
            $\mathbf{x}_{(t+1)} = clip_{\mathbf{x}, \epsilon}(\mathbf{x}_t + \alpha \cdot sign(\nabla_\mathbf{x} U(\mathbf{x}_t,F,p,T))) $ \;
            \If{$arg\,max ( F(\mathbf{x}_{t+1}) ) \neq initial prediction$}
            {
                end while\;
            }
          }
          return $\mathbf{x}_{t+1}$
        \caption{Algorithm for BIM A (Uncertainty based) \\ $\mathbf{x}$ is the benign image, $F$ is the model function learnt during training, $p$ is the dropout ratio used in dropout layers, $T$ is the number of MC dropout samples in model training mode, $N$ is the number of iterations, $\epsilon$ is the maximum allowed perturbation, $\alpha$ is the step size, }
        \label{alg:BIM-A-Uncertainty-Alg}
        \end{algorithm}

\begin{algorithm}[h]
\DontPrintSemicolon
  
  \KwIn{$\mathbf{x}  \in \mathbb{R}^m, F, p, T, N, \epsilon, \alpha$}
  \KwOut{$\mathbf{x}_{t+1}$ }
  $\mathbf{x}_0 \gets \mathbf{x}$ \;
  \While{$n < N$}
  {
    \If{$arg\,max(F(\mathbf{x}_{t+1})) \neq initial prediction$}
    {
        $condition = True$\;
    }
    \If{$condition = False$}
    {
        \tcc{update X by using below formula, F in training mode}
        $\mathbf{x}_{(t+1)} = clip_{\mathbf{x}, \epsilon}(\mathbf{x}_t + \alpha \cdot sign(\nabla_\mathbf{x} U(\mathbf{x}_t,F,p,T)))$ \;
    }
    \Else{
        \tcc{update X by using below formula, F in training mode}
        $\mathbf{x}_{(t+1)} = clip_{\mathbf{x}, \epsilon}(\mathbf{x}_t - \alpha \cdot sign(\nabla_\mathbf{x} U(\mathbf{x}_t,F,p,T)))$ \;
    }
  }
  return $\mathbf{x}_{t+1}$ 
\caption{Algorithm for BIM B (Uncertainty based) $\mathbf{x}$ is the benign image, $F$ is the model function learnt during training, $p$ is the dropout ratio used in dropout layers, $T$ is the number of MC dropout samples in model training mode, $N$ is the number of iterations, $\epsilon$ is the maximum allowed perturbation, $\alpha$ is the step size, }
\label{alg:BIM-B-Uncertainty-Alg}
\end{algorithm}

\subsubsection{Basic Iterative Attack (Hybrid Approach)}

Here, we present the pseudo-code for our Hybrid Approach in Algorithm \ref{alg:BIM-Hybrid-Alg}. Same as the previous BIM attack variants, our Hybrid Approach is also designed under $L_\infty$ norm. At each iteration, we step into both the model loss' gradient direction and model uncertainty's gradient direction. These two metrics make up for each other and yield to a better result.

        \begin{algorithm}[H]
        \DontPrintSemicolon
          \KwIn{$\mathbf{x}, F, p, T, N, \epsilon, \alpha$}
          \KwOut{$\mathbf{x}_{t+1}$ }
          $\mathbf{x}_0 \gets \mathbf{x} $ \;
          $initial prediction = arg\,max(F(\mathbf{x}_0))$ \tcc{F in evaluation mode}
          \While{$n < N$}
          {
            \tcc{update X by using below formula, F in training mode when calculating gradient of model uncertainty, F in evaluation mode when calculating gradient of model loss}
            $\mathbf{x}_{(t+1)} = clip_{\mathbf{x}, \epsilon}(\mathbf{x}_t + \alpha \cdot sign(\nabla_\mathbf{x} U(\mathbf{x}_t,F,p,T)) + \alpha \cdot sign(\nabla_\mathbf{x} \ell(\mathbf{x}_t, y_{true}))) $ \;
            \If{$arg\,max ( F(\mathbf{x}_{t+1}) ) \neq initial prediction $}
            {
                end while\;
            }
          }
          $return \,\, \mathbf{x}_{t+1}$
        \caption{Algorithm for BIM A (Hybrid Approach) $\mathbf{x}$ is the benign image, $F$ is the model function learnt during training, $p$ is the dropout ratio used in dropout layers, $T$ is the number of MC dropout samples in model training mode, $N$ is the number of iterations, $\epsilon$ is the maximum allowed perturbation, $\alpha$ is the step size, }
        \label{alg:BIM-Hybrid-Alg}
        \end{algorithm}
        

\subsection{Visualizing Gradient Path for Uncertainty Based Attacks }\label{sec:uncertainty_based_attacks}
\begin{figure}[!htbp]
    \centering
    \begin{subfigure}[b]{0.5\linewidth}
         \centering
         \captionsetup{justification=centering}
         \includegraphics[width=1\linewidth]{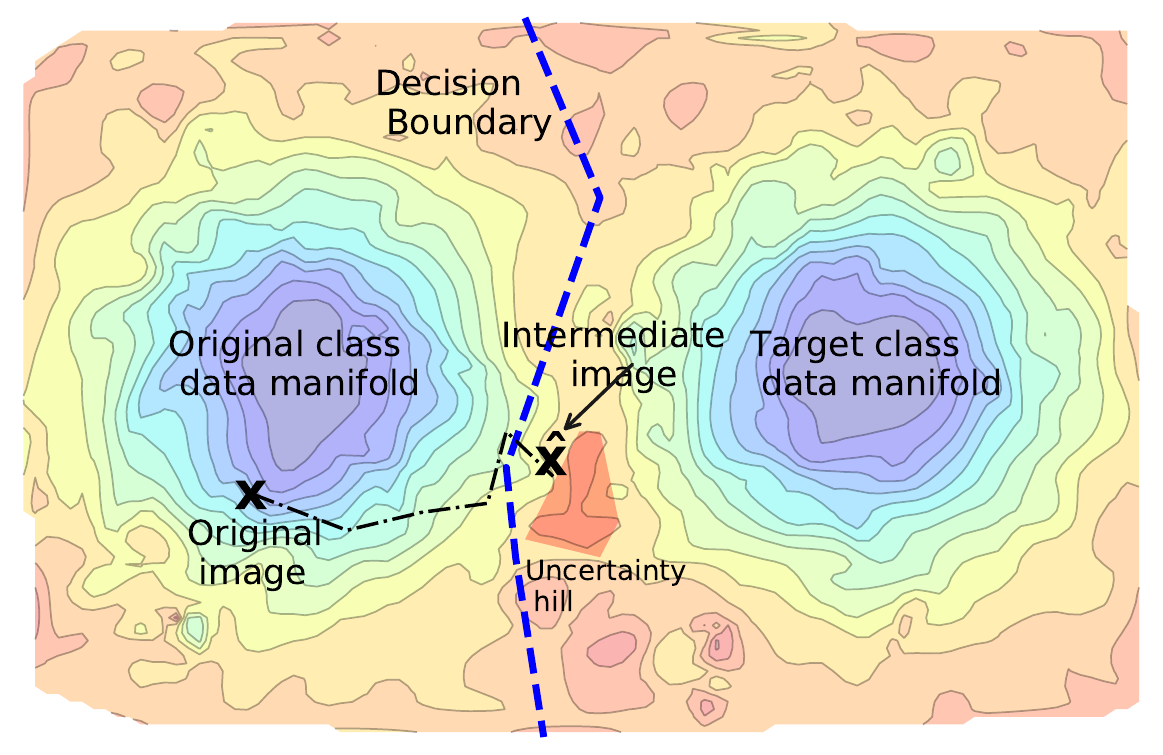}
         \caption{Type A success, Type B fail}
	    \label{fig:plot_suncattack-v1}
     \end{subfigure}\hfill
     \begin{subfigure}[b]{0.5\linewidth}
         \centering
         \captionsetup{justification=centering}
         \includegraphics[width=1\linewidth]{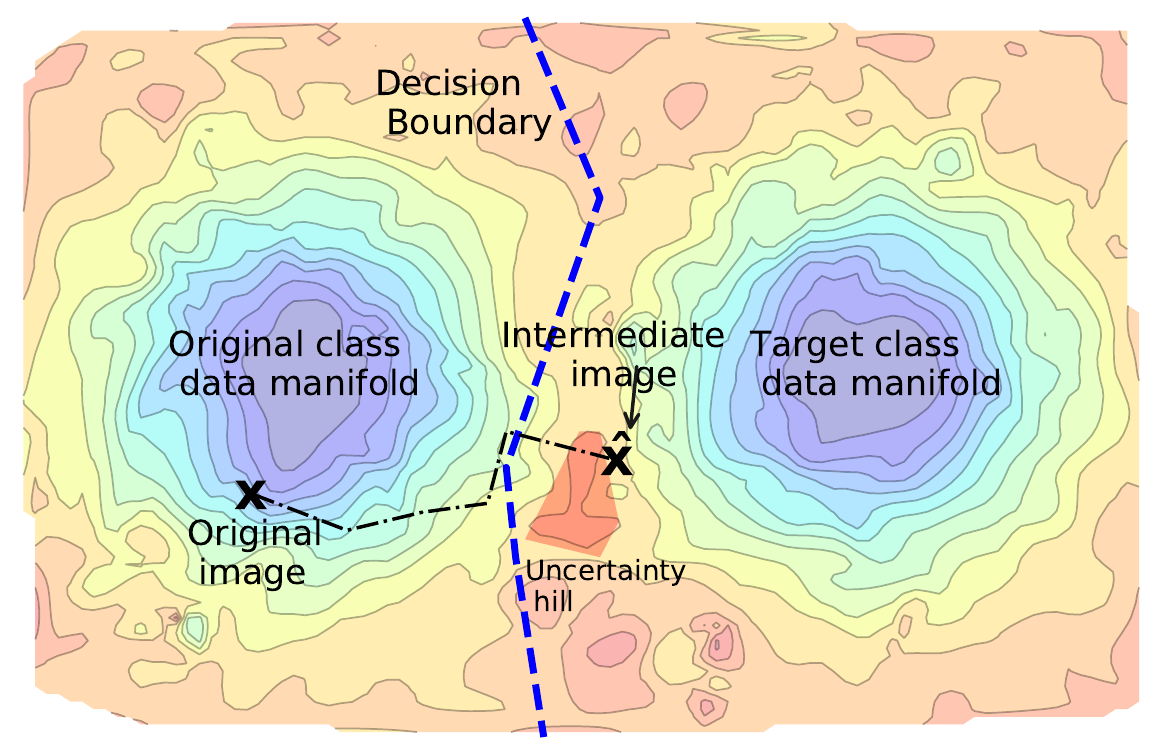}
         \caption{Type A and B success}
	    \label{fig:plot_suncattack-v2}
     \end{subfigure}\hfill
	\caption{Uncertainty gradient path}
	\label{fig:loss-surfaces-BIM}
\end{figure}

Figure \ref{fig:loss-surfaces-BIM} shows a simplified example of the gradient path for our uncertainty based BIM attack variants. In the example shown in the pictures, the low uncertainty regions are shown in blue, while the high uncertainty regions are shown in red. Figure \ref{fig:plot_suncattack-v1} shows an example of successful uncertainty based BIM attack type A. But, we would expect the uncertainty based BIM attack type B to be unsuccessful for this specific example. Because at the intermediate iteration where we passed the decision boundary from source to target class, we are at the left side of the uncertainty hill. When we try to decrease the uncertainty, we will perturb the image back to the original class manifold. However, for Figure \ref{fig:plot_suncattack-v2}, we would expect both uncertainty based BIM attack type A and B would be successful. Because uncertainty hill is at the right of the decision boundary and at the intermediate iteration, we are at the right the uncertainty hill.  

\begin{figure}[!htbp]
    \centering
    \begin{subfigure}[b]{0.48\linewidth}
         \centering
         \includegraphics[width=1.0\linewidth]{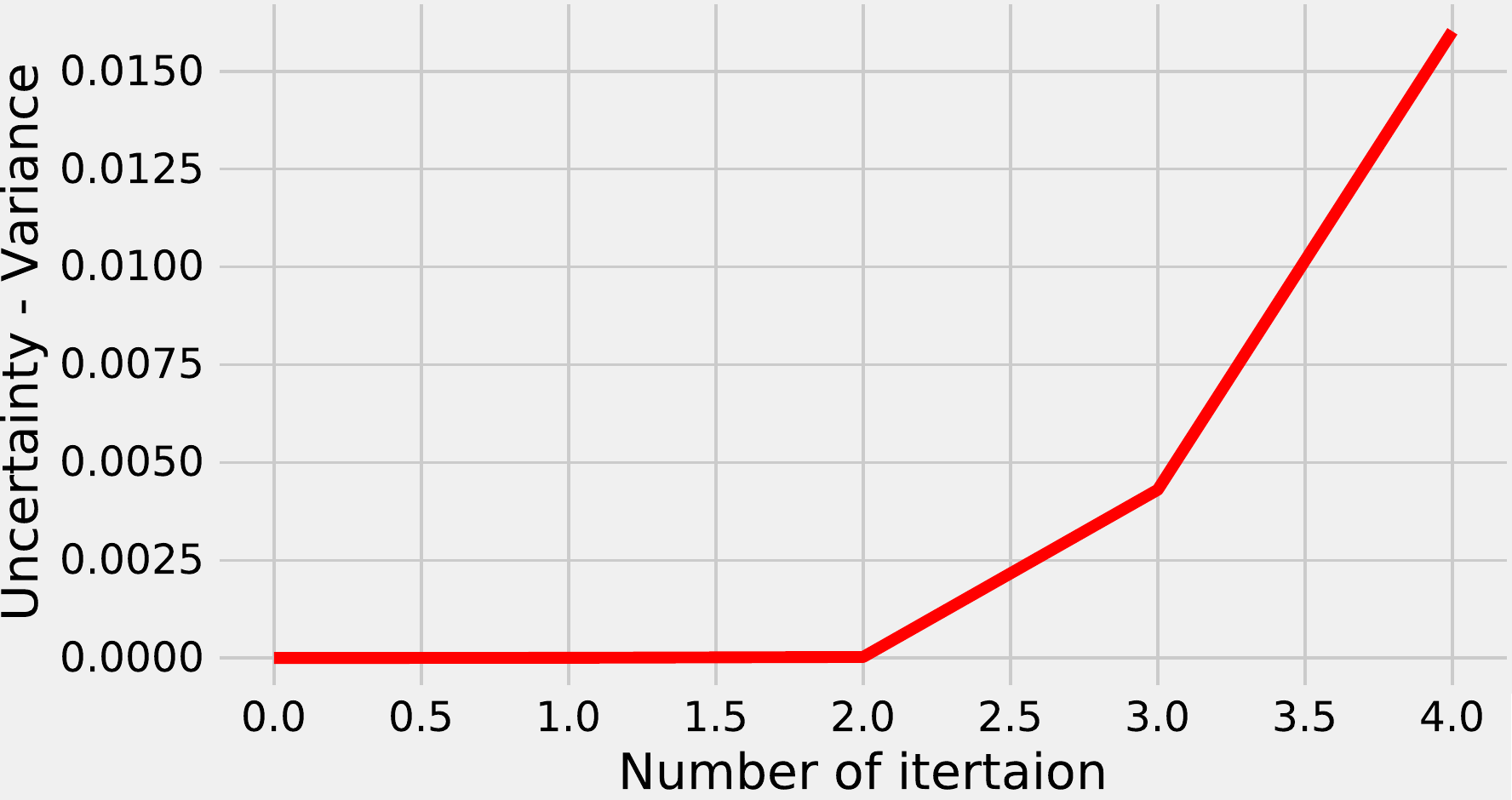}
         \caption{BIM-A (Uncertainty Based)}
	    \label{fig:variance_bim_a_uncertainty_based}
     \end{subfigure}\hfill
     \centering
    \begin{subfigure}[b]{0.48\linewidth}
         \centering
         \includegraphics[width=1.0\linewidth]{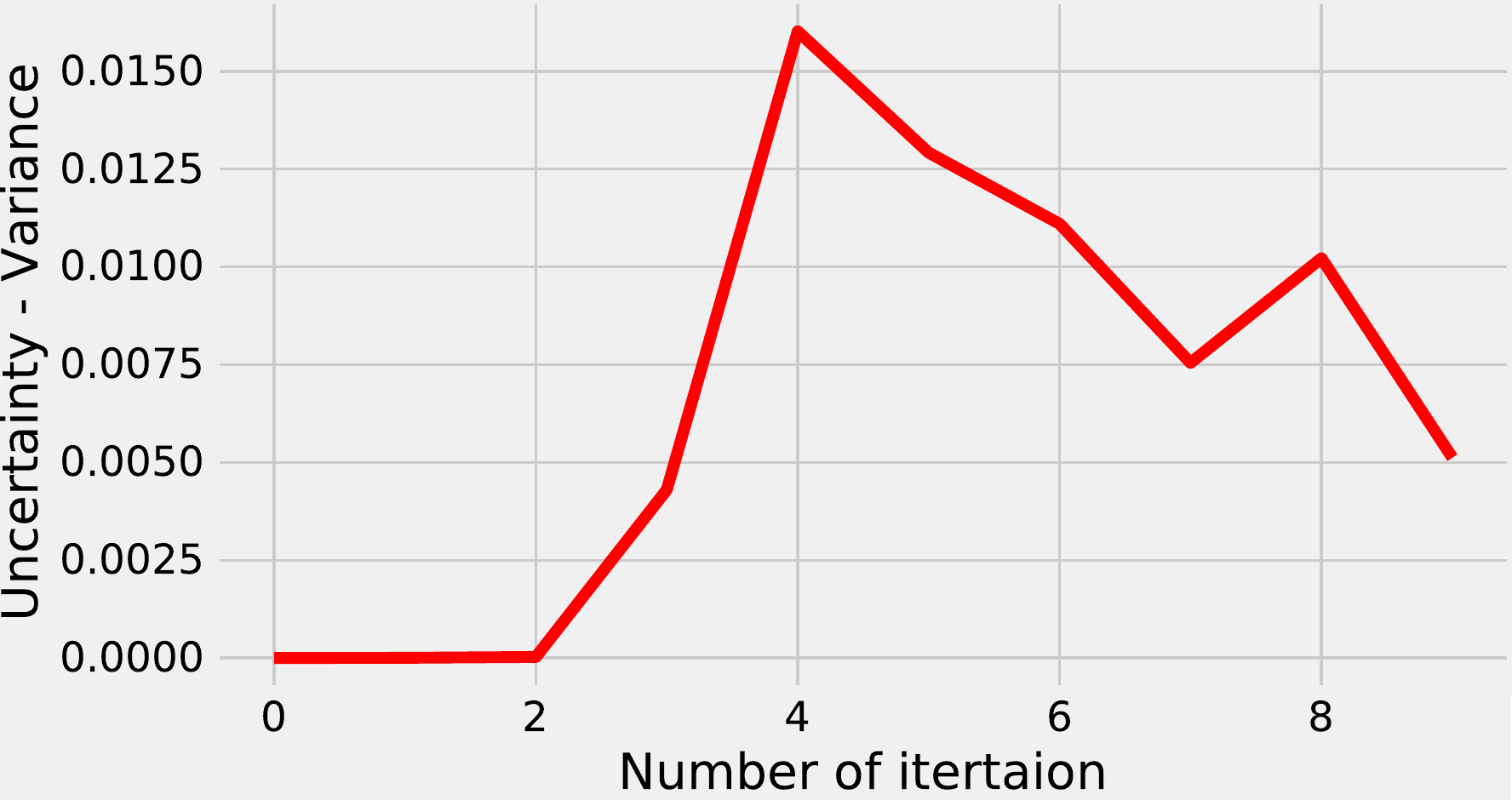}
         \caption{{BIM-B (Uncertainty Based)}}
	    \label{fig:variance_bim_b_uncertainty_based}
     \end{subfigure}\hfill
         \begin{subfigure}[b]{0.48\linewidth}
         \centering
         \includegraphics[width=1.0\linewidth]{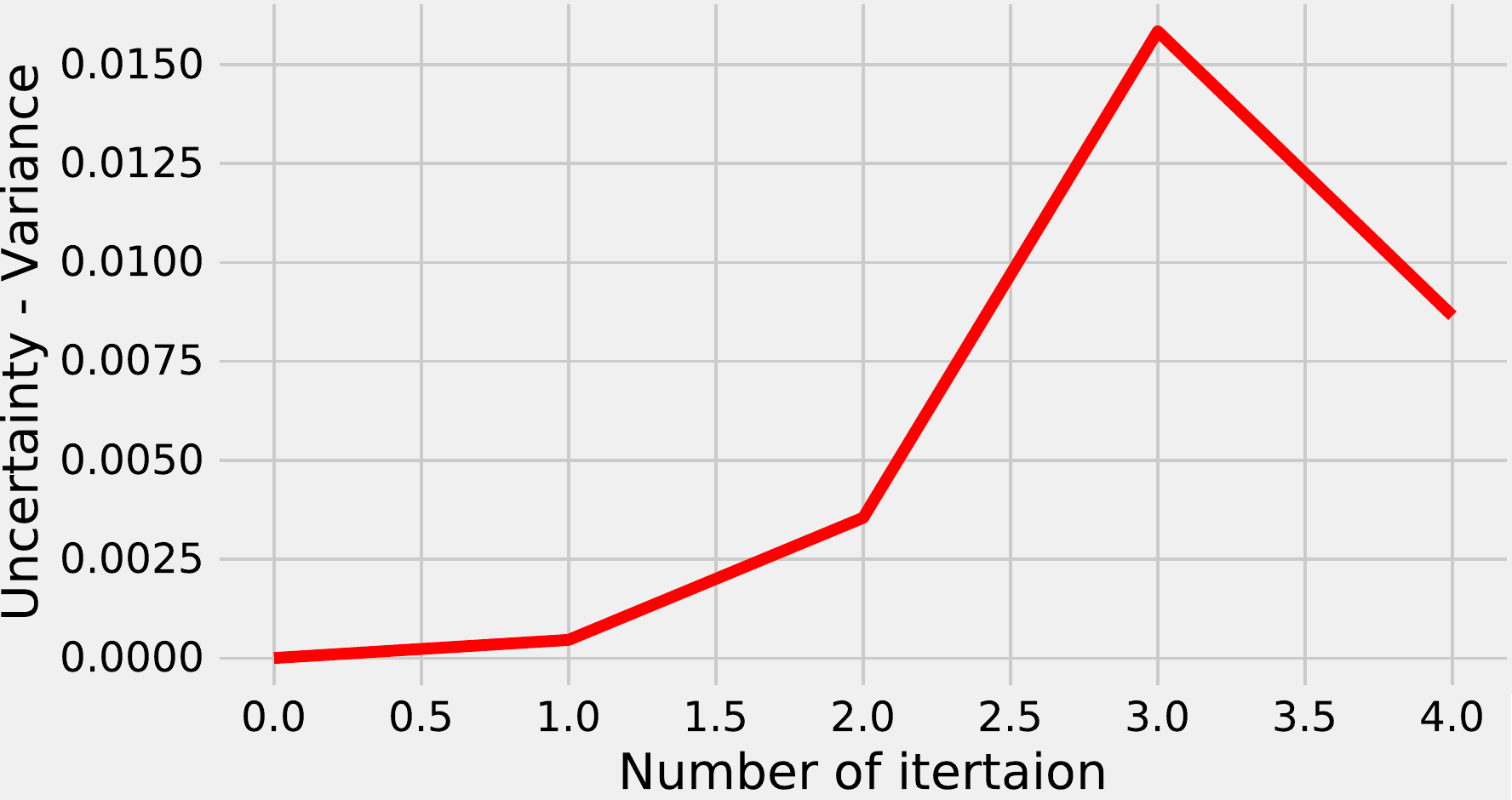}
         \caption{{BIM-A (Loss Based)}}
	    \label{fig:variance_bim_a_loss_based}
     \end{subfigure}\hfill
         \begin{subfigure}[b]{0.48\linewidth}
         \centering
         \includegraphics[width=1.0\linewidth]{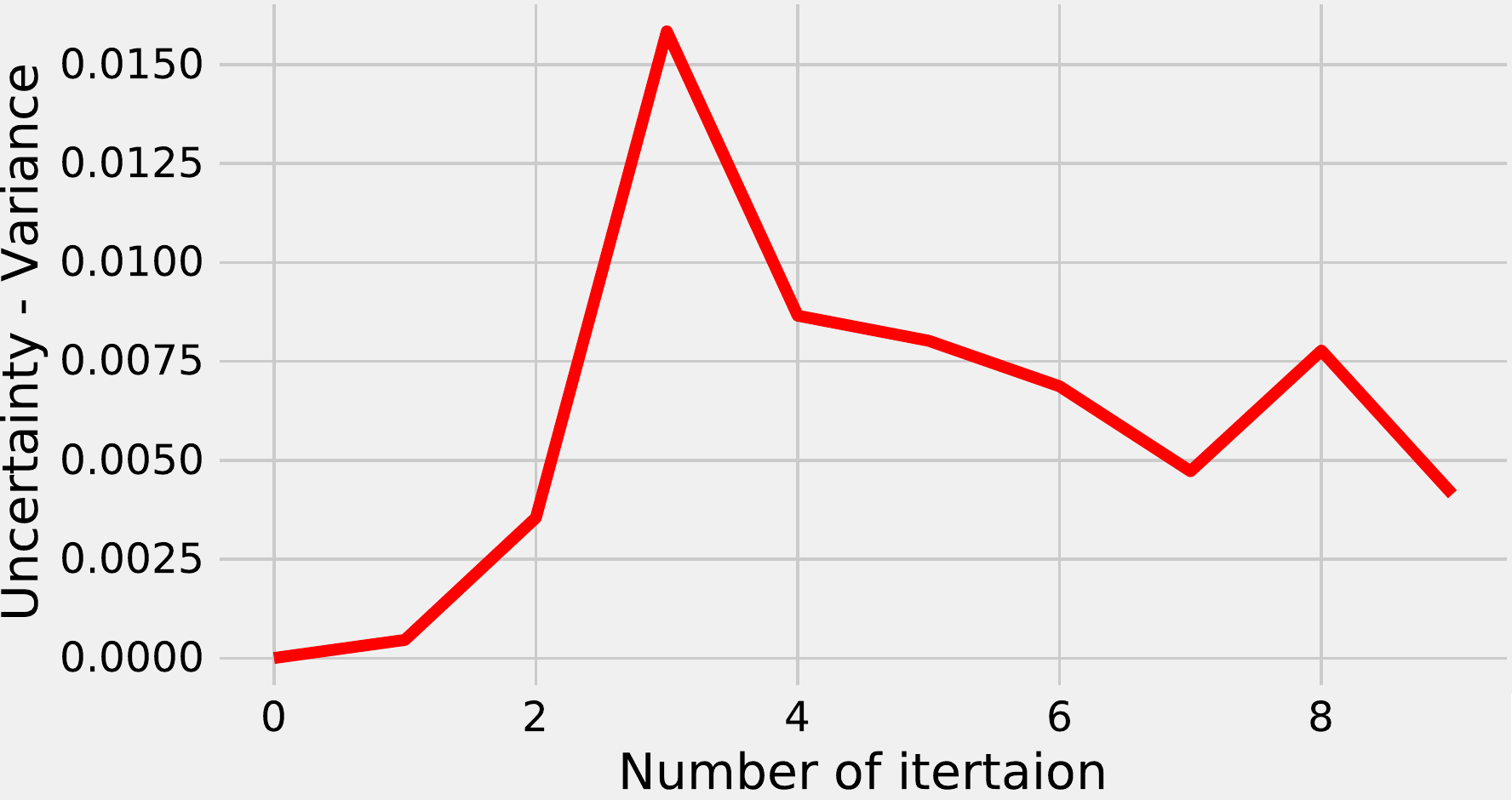}
         \caption{{BIM-B (Loss Based)}}
	    \label{fig:variance_bim_b_loss_based}
     \end{subfigure}\hfill
         \begin{subfigure}[b]{0.48\linewidth}
         \centering
         \includegraphics[width=1.0\linewidth]{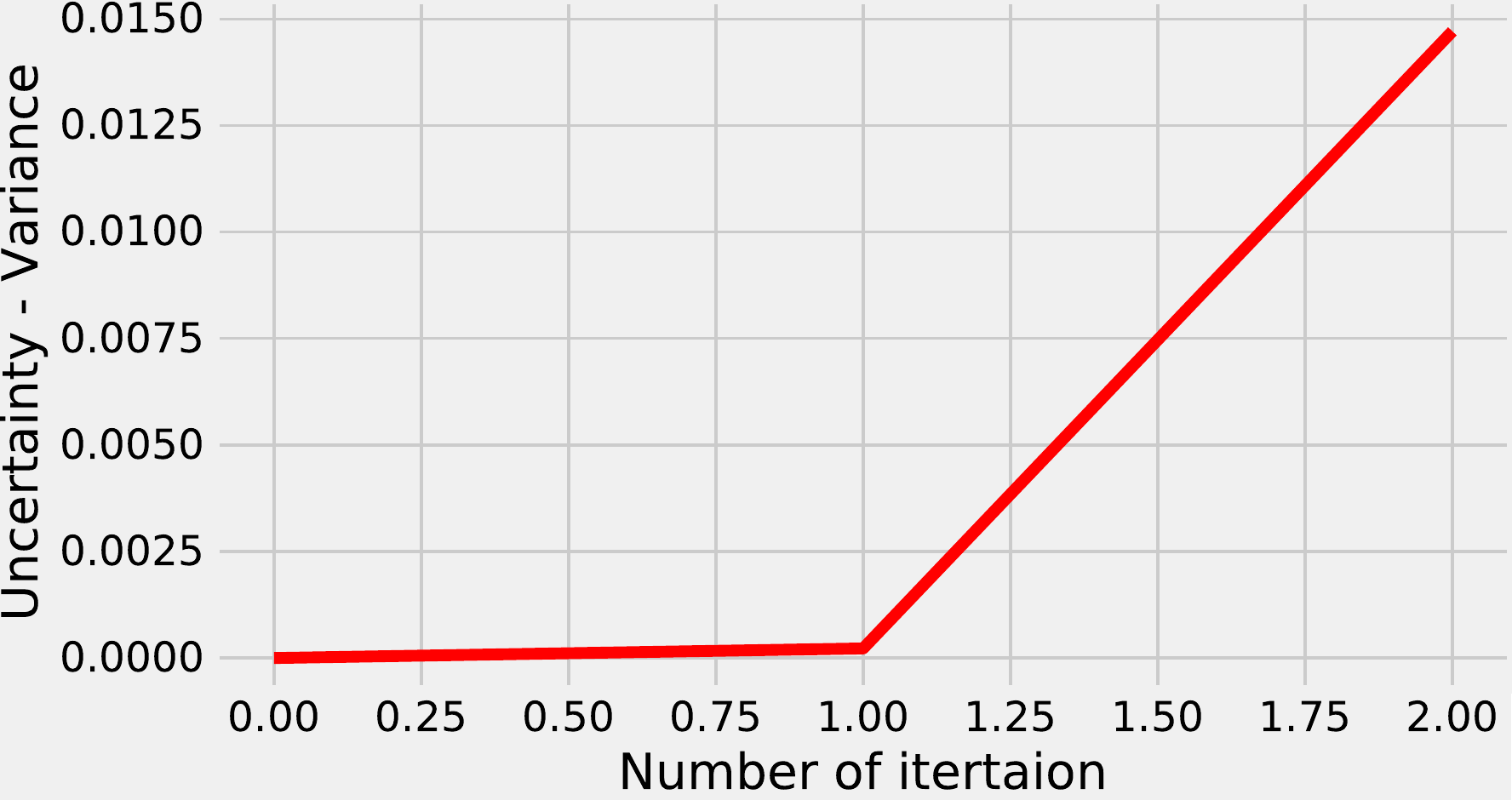}
         \caption{{BIM-Enhanced (Hybrid Approach)}}
	    \label{fig:variance_bim_enhanced_hybrid_approach}
     \end{subfigure}\hfill
    \caption{Change of uncertainty values during different BIM attack variants}
	\label{fig:variances}
\end{figure}

\subsection{Visualizing Uncertainty Under Different Attack Variants }\label{sec:attack_variants}
Figure \ref{fig:variances} shows the change in our quantified uncertainty values of the model during different BIM attack variants. In this example, we apply all of the attack variants to the 23\textsuperscript{th} test sample from MNIST (Digit) dataset. The original label of the input image was 6. For type A and B of both loss and uncertainty based attacks, we observed that at the 4\textsuperscript{th}, iteration, the attack is successful and the input image was misclassified as 4. In Figure \ref{fig:variance_bim_a_uncertainty_based},  we stop the iteration as soon as we succeeded in fooling the model, whereas in Figure \ref{fig:variance_bim_b_uncertainty_based}, we continue to perturb the image, but this time in a direction which minimize the uncertainty. After the last iteration, the predicted label was still 4 and the uncertainty level decreased compared to the time of misclassification. For this sample, our uncertainty based BIM attack type B was successful, because, when we pass the decision boundary as we try to maximize model uncertainty, we also go beyond the point where there is the maximum uncertainty. One last important point to mention is that, when we apply the hybrid approach where we utilize both loss and uncertainty, we could successfully fool the model after 2\textsuperscript{nd} iteration, which is much faster. This also proves our assumption that the hybrid approach is more effective than the others.  

\subsection{Capability of the Attacker}
We assumed that the attacker's primary purpose is to evade the model by applying a carefully crafted perturbation to the input data.  In a real-world scenario, this white-box setting is the most desired choice for an adversary that does not take the risks of being caught in a trap. The problem is that it requires the attacker to access the model from outside to generate adversarial examples. After manipulating the input data, the attacker can exploit the model's vulnerabilities in the same manner as in an adversary's sandbox environment. The classification model predicts the adversarial instances when the attacker can convert some class labels as another class label (i.e. wrong prediction).

However, to prevent this manipulation from being easily noticed by the human eye, the attacker must solve an optimization problem to decide which regions in the input data must be changed. By solving this optimization problem using one of the available attack methods   \cite{goodfellow2015explaining,kurakin2017adversarial,madry2019deep,8965459}, the attacker aims to reduce the classification performance on the adversarial data as much as possible. In this study, to limit the maximum allowed perturbation that is allowed for the attacker, we used $l_\infty$ norm which is the maximum pixel difference limit between original and adversarial image.

\section{Results}
\label{ch:results}

\subsection{Experimental Setup}\label{sec:experimental-setup}
We trained our CNN models for the MNIST (Digit) \cite{lecun-mnisthandwrittendigit-2010}  and MNIST (Fashion) \cite{xiao2017fashionmnist} datasets, and we achieved accuracy rates of 99.05 \% and 91.15 \% respectively. The model architectures are given in Table \ref{tab:cnn_model_arch_digit} and the hyperparameters selected in Table \ref{tab:cnn_model_params}. For the CIFAR10 dataset \cite{cifar10}, we used a pretrained VGG-A (11 weight layers) model \cite{simonyan2015deep} and then apply transfer learning by freezing the convolution layers, changing the number of neurons in output layer from 1000 to 10 and updating the weights of the dense layers only for 10 epoch. By this way, we achieved an accuracy rate of 88.78 \% on test data. Since the used pretrained VGG model was trained on \texttt{IMAGENET} dataset, we had to rescale the CIFAR10 images from $32\times32$ to $224\times224$. We also applied the same normalization procedure by normalizing all the pixels with $mean=[0.485, 0.456, 0.406]$ and $std=[0.229, 0.224, 0.225]$. For CIFAR-10 dataset, we could only use around \% 54 of the test data (randomly shuffled) in our experiments due to the limits of our lab environment. The adversarial settings that have been used throughout of our experiments is provided in Table \ref{tab:adversarial_settings}. Finally, we used $T = 50$ as the number of MC dropout samples when quantifying uncertainty.

\begin{table}[h]
    \centering
    \caption{CNN model architectures}
    \label{tab:cnn_model_arch_digit}
    \begin{tabular}{|c||c|c|}
        \hline
        \textbf{Dataset} & \textbf{Layer Type} &  \textbf{Layer Information}\\
        \hline \hline
        \multirow{8}{*}{MNIST (Digit)} & Convolution (padding:1) + ReLU & $3 \times 3 \times 10$ \\
        & Max Pooling & $2 \times 2$ \\
        & Convolution (padding:1) + ReLU & $3 \times 3 \times 10$ \\
        & Max Pooling & $2 \times 2$ \\
        & Convolution (padding:1) + ReLU & $3 \times 3 \times 20$ \\
        & Dropout & p : 0.25 \\
        & Convolution (padding:1) + ReLU & $3 \times 3 \times 20$ \\
        & Dropout & p : 0.25 \\
        & Fully Connected + ReLU & 980 \\
        & Fully Connected + ReLU & 100 \\
        & Output Layer & 10 \\ 
        \hline \hline
        \multirow{10}{*}{MNIST (Fashion)} & Convolution (Padding = 1) + ReLU & $3 \times 3 \times 32$ \\
        & Max Pooling & $2 \times 2$ \\
        & Convolution (Padding = 1) + ReLU & $3 \times 3 \times 32$ \\
        & Max Pooling & $2 \times 2$ \\
        & Convolution (Padding = 1) + ReLU & $3 \times 3 \times 64$ \\
        & Dropout & p : 0.5 \\
        & Convolution (Padding = 1) + ReLU & $3 \times 3 \times 64$ \\
        & Dropout & p : 0.5 \\
        & Fully Connected + ReLU & 3136 \\
        & Fully Connected + ReLU & 600 \\
        & Fully Connected + ReLU & 120 \\
        & Output Layer & 10 \\ 
        \hline
    \end{tabular}
\end{table}

\begin{table}[h]
    \centering
    \caption{CNN model parameters}
    \label{tab:cnn_model_params}
    \begin{tabular}{c|c|c} 
        \hline
        \textbf{Parameter} & \textbf{CNN Model for MNIST (Digit)} & \textbf{CNN Model for MNIST (Fashion)}\\
        \hline \hline
        Optimizer & SGD & Adam\\
        Learning rate & 0.01 & 0.001\\
        Batch Size & 100 & 64\\
        Dropout Ratio & 0.25 & 0.25\\
        Epochs & 7 & 10\\
        \hline
    \end{tabular}
\end{table}

\begin{table}[h]
    \centering
    \caption{Adversarial settings of our experiments: $\alpha$, i respectively denote the step-size and the number of attack steps for a perturbation budget $\epsilon$}
    \label{tab:adversarial_settings}
    \begin{tabular}{c|c|c} 
        \hline
        \textbf{Attack} & \textbf{Parameters} & \textbf{$l_p$ norm}\\
        \hline \hline
        FGSM & i = 1 & $l_\infty$\\
        BIM & $\alpha$ = $\epsilon$ $\cdot$ 0.2, i = 10 (for MNIST) & $l_\infty$ \\
        BIM & $\alpha$ = $\epsilon$ $\cdot$ 0.3, i = 5 (for CIFAR-10) & $l_\infty$ \\
        \hline
    \end{tabular}
\end{table}

\subsection{Experimental Results}\label{sec:experimental-results}

During our experiments, we only perturbed the test samples which were correctly classified by our
models in their original states. Because, it is obvious that an intruder would have no reason to
perturb samples that are already classified wrongly.

\begin{table}[h]
\centering
\caption{Attack success rates on different datasets }
     \label{tab:attack-success-rates}
    \begin{tabular}{c|c|c|c} 
        \hline
         & MNIST (Digit)  $\epsilon$ = 0.15 & MNIST (Fashion)  $\epsilon$ = 0.05 & CIFAR10  $\epsilon$ = 0.2/255 \\
        \hline
                FGSM (Loss based)  & \% 47.34  & \%59.40 & \%70.83\\
       FGSM (Uncertainty based)& \% 47.80  & \%62.73 & \%69.76\\
               BIM-A (Loss based) & \% 82.59 & \% 82.86 & \%88.06\\
       BIM-A (Uncertainty based)  & \% 75.68  & \% 84.71 & \%85.57\\
        BIM-B (Uncertainty based)  & \% 65.13  & \% 71.60 & \%81.05\\
       BIM-A (Hybrid Approach)  & \% 85.40  & \% 89.92 & \%90.03\\
        \hline
    \end{tabular}
\end{table}

\begin{figure}
    \centering
    \includegraphics[width=1.0\linewidth]{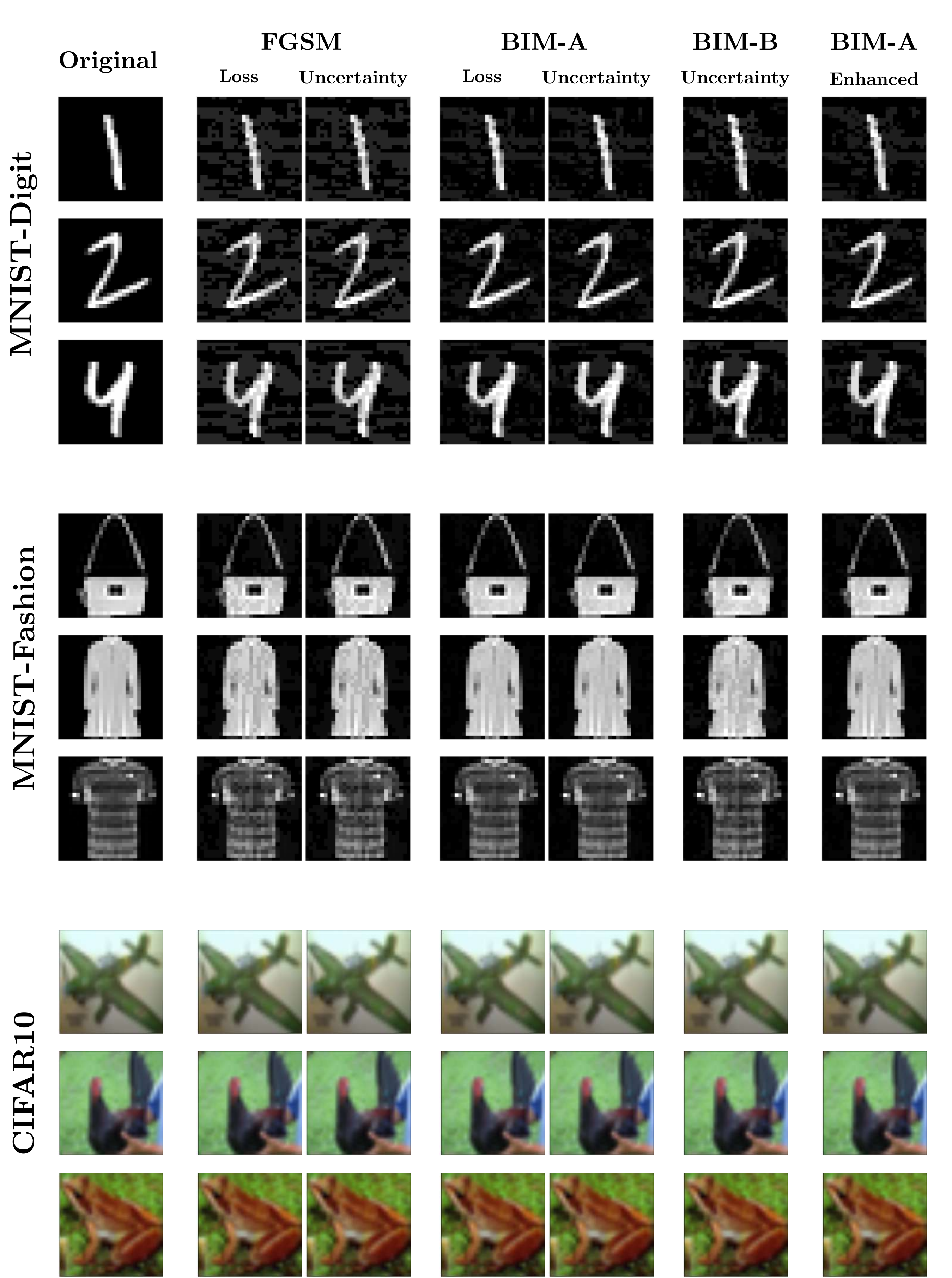}
    \caption{Some example images from MNIST(Digit), MNIST(Fashion) and CIFAR-10. The original image is shown in the left-most column and adversarial samples crafted based on different methods are on the other columns.}
    \label{fig:overall_plott.pdf}
\end{figure}

The results show that our Hybrid Approach of using both model's loss and uncertainty results into the best performance. Success rates of pure loss based and pure uncertainty based attacks are similar to each other. We also observe that the success rates for uncertainty based attack types A and B are different. We argue that the point of global maximum for uncertainty metric for any class is not on the model's decision boundary as in the case of model loss. Instead, the point where the uncertainty is maximum can be beyond the decision boundary. Therefore, during the gradient-based search, it may be possible for us to pass the decision boundary but still not reaching the peak value for uncertainty. And when we start to decrease the uncertainty after passing the decision boundary (fooling the model), it may be possible to go back to the original class. However, this is not the case in loss based approaches. Since we are trying to maximize the loss based on a reference class, we always see an increasing trend during the journey of gradient descent approach of loss maximization. Figure \ref{fig:overall_plott.pdf} shows some examples of adversarial samples crafted using different methods mentioned in this study.

\section{Conclusion}
\label{ch:conclusion}
In this study, we proposed a new attack algorithm by perturbing the input in a direction that maximizes the model's epistemic uncertainty instead of the model's loss. We observed almost similar performances compared to loss based approaches. We also introduced a new concept for finding better points resulting in higher loss values within a specified $L_p$ norm interval to craft adversarial samples. For this, we used a hybrid approach and stepped into both the loss' and uncertainty's gradient directions in each gradient descent step.  We showed that the attack success rates are higher when we utilize this approach.


The aim for this study was not to propose the most powerful attack to date, but instead, we aimed at showing that there  exists some other powerful metrics, different from model's loss, that can be exploited to craft adversarial examples. Besides, relying just on the trained model is actually not a good idea all the time as it is just an approximation to the best predictor and epistemic uncertainty information can be very useful in cases where the model is misleading. We also showed that the combined usage of uncertainty and loss yields better performance.

As future work, we will be investigating the usage of aleatoric uncertainty for adversarial attack and defense purposes, especially in detecting adversarial samples.

\bibliographystyle{unsrt} 
\bibliography{references}

\end{document}